\documentclass[english,11pt]{article}
\usepackage[T1]{fontenc}
\usepackage[latin1]{inputenc}
\usepackage{subfig}
\usepackage{xcolor}
\usepackage{float}
\usepackage{graphicx}
\usepackage{setspace}
\usepackage{amsmath, amsthm, amssymb, amsfonts, mathtools}
\usepackage{placeins}
\usepackage{babel}
\usepackage{acronym}
\usepackage{units}
\usepackage{url}
\usepackage{authblk}
\usepackage[square]{natbib}
\usepackage[margin=1.0in]{geometry}
\graphicspath{{img/}{.}}

\renewcommand{\refeq}[1]{{Eq.~(\ref{#1})}}

\newcommand{\reffig}[1]{{Fig.~\ref{#1}}}
\newcommand{\reftab}[1]{{Tab.~\ref{#1}}}

\newcommand{\refsec}[1]{{Sec.~\ref{#1}}}
\renewcommand{\cite}[1]{{\color{black!40}\citep{#1}}}

\makeatother
\begin{document}

\acrodef{AC}[AC]{Arrenhius \& Current}
\acrodef{AER}[AER]{Address Event Representation}
\acrodef{AEX}[AEX]{AER EXtension board}
\acrodef{AMDA}[AMDA]{``AER Motherboard with D/A converters''}
\acrodef{API}[API]{Application Programming Interface}
\acrodef{BM}[BM]{Boltzmann Machine}
\acrodef{CAVIAR}[CAVIAR]{Convolution AER Vision Architecture for Real-Time}
\acrodef{CCN}[CCN]{Cooperative and Competitive Network}
\acrodef{CD}[CD]{Contrastive Divergence}
\acrodef{eCD}[eCD]{event-driven Contrastive Divergence}
\acrodef{CMOS}[CMOS]{Complementary Metal--Oxide--Semiconductor}
\acrodef{COTS}[COTS]{Commercial Off-The-Shelf}
\acrodef{CPU}[CPU]{Central Processing Unit}
\acrodef{CV}[CV]{Coefficient of Variation}
\acrodef{CV}[CV]{Coefficient of Variation}
\acrodef{DAC}[DAC]{Digital--to--Analog}
\acrodef{DBN}[DBN]{Deep Belief Network}
\acrodef{DFA}[DFA]{Deterministic Finite Automaton}
\acrodef{DFA}[DFA]{Deterministic Finite Automaton}
\acrodef{divmod3}[DIVMOD3]{divisibility of a number by 3}
\acrodef{DPE}[DPE]{Dynamic Parameter Estimation}
\acrodef{DPI}[DPI]{Differential-Pair Integrator}
\acrodef{DSP}[DSP]{Digital Signal Processor}
\acrodef{DVS}[DVS]{Dynamic Vision Sensor}
\acrodef{EDVAC}[EDVAC]{Electronic Discrete Variable Automatic Computer}
\acrodef{EIF}[EI\&F]{Exponential Integrate \& Fire}
\acrodef{EIN}[EIN]{Excitatory--Inhibitory Network}
\acrodef{EPSC}[EPSC]{Excitatory Post-Synaptic Current}
\acrodef{EPSP}[EPSP]{Excitatory Post--Synaptic Potential}
\acrodef{FPGA}[FPGA]{Field Programmable Gate Array}
\acrodef{FSM}[FSM]{Finite State Machine}
\acrodef{GPU}[GPU]{Graphical Processing Unit}
\acrodef{HAL}[HAL]{Hardware Abstraction Layer}
\acrodef{HH}[H\&H]{Hodgkin \& Huxley}
\acrodef{HMM}[HMM]{Hidden Markov Model}
\acrodef{HW}[HW]{Hardware}
\acrodef{hWTA}[hWTA]{Hard Winner--Take--All}
\acrodef{IF2DWTA}[IF2DWTA]{Integrate \& Fire 2--Dimensional WTA}
\acrodef{IF}[I\&F]{Integrate \& Fire}
\acrodef{IFSLWTA}[IFSLWTA]{Integrate \& Fire Stop Learning WTA}
\acrodef{INCF}[INCF]{International Neuroinformatics Coordinating Facility}
\acrodef{INI}[INI]{Institute of Neuroinformatics}
\acrodef{IO}[IO]{Input-Output}
\acrodef{IPSC}[IPSC]{Inhibitory Post-Synaptic Current}
\acrodef{ISI}[ISI]{Inter--Spike Interval}
\acrodef{JFLAP}[JFLAP]{Java - Formal Languages and Automata Package}
\acrodef{LIF}[LI\&F]{Linear Integrate \& Fire}
\acrodef{LSM}[LSM]{Liquid State Machine}
\acrodef{LTD}[LTD]{Long-Term Depression}
\acrodef{LTI}[LTI]{Linear Time-Invariant}
\acrodef{LTP}[LTP]{Long-Term Potentiation}
\acrodef{LTU}[LTU]{Linear Threshold Unit}
\acrodef{MCMC}{Markov Chain Monte Carlo}
\acrodef{NHML}[NHML]{Neuromorphic Hardware Mark-up Language}
\acrodef{NMDA}[NMDA]{NMDA}
\acrodef{NME}[NE]{Neuromorphic Engineering}
\acrodef{PCB}[PCB]{Printed Circuit Board}
\acrodef{PRC}[PRC]{Phase Response Curve}
\acrodef{PSC}[PSC]{Post-Synaptic Current}
\acrodef{PSP}[PSP]{Post--Synaptic Potential}
\acrodef{RI}[KL]{Kullback-Leibler}
\acrodef{RRAM}[RRAM]{Resistive Random-Access Memory}
\acrodef{RBM}[RBM]{Restricted Boltzmann Machine}
\acrodef{ROC}[ROC]{Receiver Operator Characteristic}
\acrodef{SAC}[SAC]{Selective Attention Chip}
\acrodef{SSM}[S2M]{Synaptic Sampling Machine}
\acrodef{dSSM}[S2M]{Synaptic Sampling Machine}
\acrodef{S3M}[spiking S2M]{Spiking S2M}
\acrodef{SCD}[SCD]{Spike-Based Contrastive Divergence}
\acrodef{SCX}[SCX]{Silicon CorteX}
\acrodef{STDP}[STDP]{Spike Timimg Dependent Plasticity}
\acrodef{SW}[SW]{Software}
\acrodef{sWTA}[SWTA]{Soft Winner--Take--All}
\acrodef{VHDL}[VHDL]{VHSIC Hardware Description Language}
\acrodef{VLSI}[VLSI]{Very  Large  Scale  Integration}
\acrodef{WTA}[WTA]{Winner--Take--All}
\acrodef{XML}[XML]{eXtensible Mark-up Language}
\acused{S3M}
\title{Stochastic Synapses Enable Efficient Brain-Inspired Learning Machines}
\date{}
\author[1]{Emre Neftci}
\author[2]{Bruno Pedroni}
\author[3]{Siddharth Joshi}
\author[4]{Maruan Al-Shedivat}
\author[2]{Gert Cauwenberghs}
\affil[1]{Department of Cognitive Sciences, UC Irvine}
\affil[2]{Department of Bioengineering, UC San Diego}
\affil[3]{Electrical and Computer Engineering Department, UC San Diego}
\affil[4]{Machine Learning Department, Carnegie Mellon University}

\maketitle
\begin{abstract}
Recent studies have shown that synaptic unreliability is a robust and sufficient mechanism for inducing the stochasticity observed in cortex.
Here, we introduce Synaptic Sampling Machines, a class of neural network models that uses synaptic stochasticity as a means to Monte Carlo sampling and unsupervised learning.
Similar to the original formulation of Boltzmann machines, these models can be viewed as a stochastic counterpart of Hopfield networks, but where stochasticity is induced by a random mask over the connections. Synaptic stochasticity plays the dual role of an efficient mechanism for sampling, and a regularizer during learning akin to DropConnect.
A local synaptic plasticity rule implementing an event-driven form of contrastive divergence enables the learning of generative models in an on-line fashion.
Synaptic sampling machines perform equally well using discrete-timed artificial units (as in Hopfield networks) or continuous-timed leaky integrate \& fire neurons.
The learned representations are remarkably sparse and robust to reductions in bit precision and synapse pruning: removal of more than 75\% of the weakest connections followed by cursory re-learning causes a negligible performance loss on benchmark classification tasks.
The spiking neuron-based synaptic sampling machines outperform existing spike-based unsupervised learners, while potentially offering substantial advantages in terms of power and complexity, and are thus promising models for on-line learning in brain-inspired hardware.
\end{abstract}

\section{Introduction}
The brain's cognitive power emerges from a collective form of computation extending over very large ensembles of sluggish, imprecise, and unreliable components.
This realization spurred scientists and engineers to explore the remarkable mechanisms underlying biological cognitive computing by reverse engineering  the brain in ``neuromorphic'' silicon, providing a means to validate hypotheses on neural structure and function through ``analysis by synthesis''.
Successful realizations of large-scale neuromorphic hardware \cite{Indiveri_etal11_neur-sili,Park_etal14_65k--73-m,Giulioni_etal15_real-time,Merolla_etal14_mill-spik,Schemmel_etal10_wafe-neur,Benjamin_etal14_neur-mixe} are confronted with challenges in configuring and deploying them for solving practical tasks, as well as identifying the domains of applications where such devices could excel.
The central challenge is to devise a neural computational substrate that can efficiently perform the necessary inference and learning operations in a scalable manner using limited memory resources and limited computational primitives, while relying on temporally and spatially local information. 

Recently, one promising approach to configure neuromorphic systems for practical tasks is to take inspiration from machine learning, namely artificial and deep neural networks.
Despite the fact that neural networks were first inspired by biological neurons, the mapping of modern machine learning techniques onto neuromorphic architectures requires overcoming several conceptual barriers. 
This is because machine learning algorithms and artificial neural networks are designed to operate efficiently on digital processors, often using batch, discrete-time, iterative updates (lacking temporal locality), shared parameters and states (lacking spatial locality), and minimal interprocess communication that optimize the capabilities of GPUs or multicore processors. 

Here, we report a class of stochastic neural networks that overcomes this conceptual barrier while offering substantial improvements in terms of performance, power and complexity over existing methods for unsupervised learning in spiking neural networks.
Similarly to how Boltzmann machines were first proposed \cite{Hinton_Sejnowski86_lear-rele}, these neural networks can be viewed as a stochastic counterpart of Hopfield networks \cite{Hopfield82_neur-netw}, but where stochasticity is caused by multiplicative noise at the connections (synapses).

As in neural sampling \cite{Berkes_etal11_spon-cort,Fiser_etal10_stat-opti}, neural states (such as instantaneous firing rates or individual spikes) in our model are interpreted as Monte Carlo samples of a probability distribution.
The source of the stochasticity in neural samplers is often unspecified, or is assumed to be caused by independent, background Poisson activities \cite{Petrovici_etal13_stoc-infe,Neftci_etal14_even-cont}.
Background Poisson activity can be generated self-consistently in balanced excitatory-inhibitory networks \cite{Vreeswijk_Sompolinsky96_chao-neur}, or by using finite-size effects and neural mismatch \cite{Amit_Brunel97_dyna-recu}. 
Although a large enough neural sampling network could generate its own stochasticity in this fashion, 
the variability in the spike trains decreases strongly as the firing rate increases \cite{Fusi_Mattia99_coll-beha,Moreno-Bote14_pois-spik,Brunel00_dyna-spar}, unless there is an external source of noise or some parameters are fine-tuned.
Furthermore, when a neural sampling network generates its own noise, correlations in the background activity can play an adverse role in its performance \cite{Probst_etal15_prob-infe}.
Correlations can be mitigated by adding feed-forward connectivity between a balanced network (or other dedicated source of Poisson spike trains) and the neural sampler, but such solutions entail increased resources in neurons, connectivity and ultimately energy.
Therefore, the above techniques for generating stochasticity are not ideal for neural sampling.

On the other hand, synaptic unreliability can induce the necessary stochasticity without requiring a dedicated source of Poisson spike trains.
The unreliable transmission of synaptic vesicles in biological neurons is a well studied phenomenon \cite{Katz66_nerv-musc,Branco_Staras09_prob-neur}, and many studies suggested it as a major source of stochasticity in the brain \cite{Faisal_etal08_nois-nerv,Abbott_Regehr04_syna-comp,Yarom_Hounsgaard11_volt-fluc,Moreno-Bote14_pois-spik}. 
In the cortex, synaptic failures were argued to reduce energy consumption while maintaining the computational information transmitted by the post-synaptic neuron \cite{Levy_Baxter02_ener-neur}.
More recent work suggested synaptic sampling as a mechanism for representing uncertainty in the brain \cite{Aitchison_Latham15_syna}, and its role in synaptic plasticity and rewiring \cite{Kappel_etal15_netw-plas}.  

We show that uncertainty at the synapse is sufficient in inducing the variability necessary for probabilistic inference in brain-like circuits.
Strikingly, we find that the simplest model of synaptic unreliability, called \emph{blank-out} synapses, can vastly improve the performance of spiking neural networks in practical machine learning tasks over existing solutions, while being extremely easy to implement in hardware \cite{Goldberg_etal01_prob-syna}, and often naturally occurring in emerging memory technologies \cite{Saghi_etal15_plas-memr,Al-Shedivat_etal15_inhe-stoc,Yu_etal13_stoc-lear}.
In blank-out synapses, for each pre-synaptic spike-event, a synapse evokes a response at the post-synaptic neuron with a probability smaller than one. 
In the theory of stochastic processes, the operation of removing events from a point process with a probability $p$ is termed $p$-thinning.
Thinning a point process has the interesting property of making the process more Poisson-like. 
Along these lines, a recent study showed that spiking neural networks endowed with stochastic synapses robustly generate Poisson-like variability over a wide range of firing rates \cite{Moreno-Bote14_pois-spik}, a property observed in the cortex. 
The iterative process of adding spikes through neuronal integration and removing them through probabilistic synapses causes the spike statistics to be Poisson-like over a large range of firing rates.
A neuron subject to other types of noise, such as white noise injected to the soma, spike jitter and random
synaptic delays tends to fire more regularly for increasing firing firing rates.

\begin{figure*}
    \centering
    \includegraphics[width=.9\textwidth]{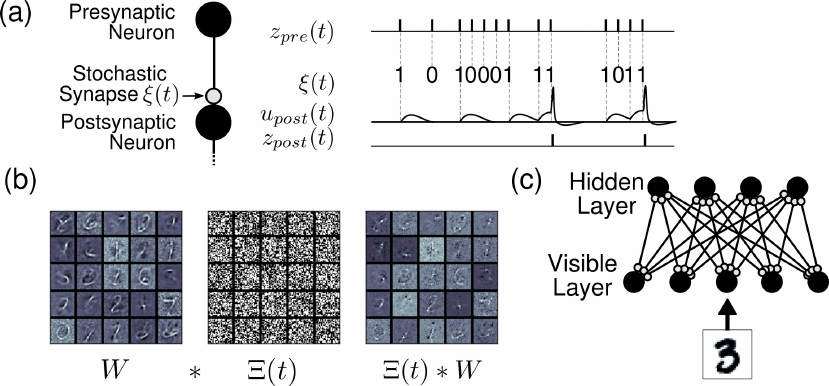}
    \caption{\label{fig:srbm} The \acf{SSM}. (a) At every occurrence of a pre-synaptic event, a pre-synaptic event is propagated to the post-synaptic neuron with probability $p$. (b) Synaptic stochasticity can be viewed as a continuous DropConnect method \cite{Wan_etal13_regu-neur} where weights are masked by a binary matrix $\Theta(t)$, where $\ast$ denotes element-wise multiplication. (c) \ac{SSM} Network architecture, consisting of a visible and a hidden layer.}
\end{figure*}

Given the strong inspiration from Boltzmann machines, and that probabilistic synapses are the main source of stochasticity for the sampling process, we name such networks \acp{SSM} (\reffig{fig:srbm}).
Synaptic noise in the \ac{SSM} is not only an efficient mechanism for implementing stochasticity in spiking neural networks but also plays the role of a regularizer during learning, akin to DropConnect \cite{Wan_etal13_regu-neur}.
This technique was used to train artificial neural networks used multiplicative noise on one layer of a feed-forward network for regularization and decorrelation. 

Compared to the neural sampler used in \cite{Neftci_etal14_even-cont}, the \ac{SSM} comes at the cost of a quantitative link between the parameters of the probability distribution and those of the spiking neural network.
In a machine learning task, this does not pose a problem since the parameters of the spiking neural network can be trained with a learning rule such as \ac{eCD}.
ECD is an on-line training algorithm for a subset of stochastic spiking neural networks \cite{Neftci_etal14_even-cont}:
The stochastic neural network produces samples from a probability distribution, and \ac{STDP} carries out the weight updates according to the Contrastive Divergence rule \cite{Hinton02_trai-prod} in an online, asynchronous fashion.

\acp{SSM} trained with \ac{eCD} significantly outperform the \ac{RBM}, reaching error rates in an MNIST hand-written digit recognition task of $4.4\%$, while using a fraction of the number of synaptic operations during learning and inference compared to our previous implementation \cite{Neftci_etal14_even-cont}.
Furthermore, the system has two appealing properties: 1) The activity in the hidden layer is very sparse, with less than 10\% of hidden neurons being active at any time; 2) The \ac{SSM} is remarkably robust to the pruning and the down-sampling of the weights: 
Upon pruning the top $80\%$ of the connections (sorted by weight values) and re-training (32k samples), we observed that the error rate remained below $5\%$. 
Overall, the \ac{SSM} outperforms existing models for unsupervised learning in spiking neural networks while potentially using a fraction of the resources, making it an excellent candidate for implementation in hardware. 

The article is structured as follows: In the Methods section, we first describe the \ac{SSM} within a discrete-time framework to gain insight into the functionality of the sampling process. 
We then describe the continuous-time, spiking version of the \ac{SSM}. 
In the Results section, we demonstrate the ability of \acp{SSM} to learn generative models of the MNIST dataset, and illustrate some of its remarkable features.

\section{Materials and Methods}

\subsection{The \ac{dSSM} as a Hopfield Network with Multiplicative Noise}
We first define the \ac{dSSM} as a modification of the original \acl{BM}.
Boltzmann machines are Hopfield networks with thermal noise added to the neurons in order to avoid overfitting and falling into spurious local minima \cite{Hinton_Sejnowski86_lear-rele}.
As in Boltzmann Machines, the \ac{dSSM} introduces a stochastic component to Hopfield networks.
But rather than units themselves being noisy, in the \ac{dSSM} the connections are noisy.

In the Hopfield network, neurons are threshold units:
\begin{equation}\label{eq:hopfield-activation}
    z_i[t] = \Theta(u_i[t]) = \begin{cases} r_{\text{on}} &\mbox{if } u_i[t] \ge 0 \\
        r_{\text{off}} & \mbox{if } u_i[t] < 0 \end{cases} \forall i,
\end{equation}
and connected recurrently through symmetric weights $w_{ij}$ = $w_{ji}$, where $r_{\text{on}}>r_{\text{off}}$ are two numbers indicating the values of the on and off states, respectively.

For simplicity, we chose blank-out synapses as the model of multiplicative noise. 
With blank-out synapses, the synaptic input to Hopfield unit $i$ is:
\begin{equation}
    u_i[t+1] = \sum_{j=1}^N \xi_{ij}^p[t] z_j[t] w_{ij} + b_i,
\end{equation}
where $\xi_{ij}^p[t] \in \{ 0,1\}$ is a Bernoulli process with probability $p$ and $b_i$ is the bias.
This corresponds to a weighted sum of Bernoulli variables.
Since the $\xi_{ij}^p[t]$ are independent, for large $N$ and $p$ not close to $0$ or $1$, we can approximate this sum with a Gaussian with mean $\mu_i[t] = b_i + \sum_j p w_{ij} z_j[t]$ and variance $\sigma_i^2[t] = p(1-p)\sum_j (w_{ij} z_j[t])^2$\footnote{The Gaussian approximation of a sum of Bernouilli variables is deemed valid when $Np,N(1-p)>5$, where $N$ is here the number of draws.}.
So,
\begin{equation}
    u_i[t+1] = \mu_i[t]  + \sigma_i[t] \eta_i [t],
\end{equation}
where $\eta \sim N(0,1)$. 
In other terms $u_i[t+1] \sim N(\mu_i[t], \sigma_i^2[t])$.
Since $P(z_i [t+1] = 1| \mathbf{z} [t]) = P(u_i [t+1] \ge 0| \mathbf{z} [t])$, the probability that unit $i$ is active given the network state is equal to one minus the cumulative distribution function of $u_i$:
\[
    P(u_i [t+1] \ge 0| \mathbf{z} [t]) = \frac12 \left( 1 + \operatorname{erf}\left( \frac{\mu_i[t]}{\sigma_i[t]\sqrt{2}}\right) \right),
\]
where ``$\mathrm{erf}$'' stands for the error function.
Plugging back to \refeq{eq:hopfield-activation} gives:

\begin{equation}\label{eq:ensemble_hopfield}
    \begin{split}
    P(z_i[t+1] = 1| \mathbf{z}[t]) &= P(u_i [t+1] \ge 0| \mathbf{z}[t])\\
    &= \frac12 \left(1 + \mathrm{erf}\left(  \beta    \frac{\sum_j w_{ij} z_j[t] + \frac{b_i}{p}}{\sqrt{\sum_j w_{ij}^2 z_j[t]^2}} \right)\right), \\
    \beta                &=  \sqrt{\frac{ p }{2(1-p) } }.
    \end{split}
\end{equation}

This activation function is the S2M-equivalent of the activation function as defined for neurons of the Boltzmann machines (\emph{i.e.} the logistic function).
However, four properties distinguish the S2M from the Boltzmann machine: 1) The activation function is a (shifted) error function rather than the logistic function, 2) the synaptic contributions to each neuron $i$ are normalized such that Euclidian norm of the vector $\left( w_{i1} z_1, \dots, w_{iN} z_N \right)^\top$ is $1$. Interestingly, this means that according to \refeq{eq:ensemble_hopfield}, a rescaling of the weights has the same effect of scaling the bias. If the biases are zero, the rescaling has no effect; 3) $\beta$, which depends on the probability of the synaptic blank-out noise and the network state, plays the role of thermal noise in the Boltzmann machine; and 4) In general, despite that the connectivity matrix is symmetric, the interactions in the \ac{dSSM} are asymmetric because the effective slope of the activation function depends on the normalizing factor of each neuron.

\begin{figure}[h]
    \centering
    \includegraphics[width=0.4\textwidth]{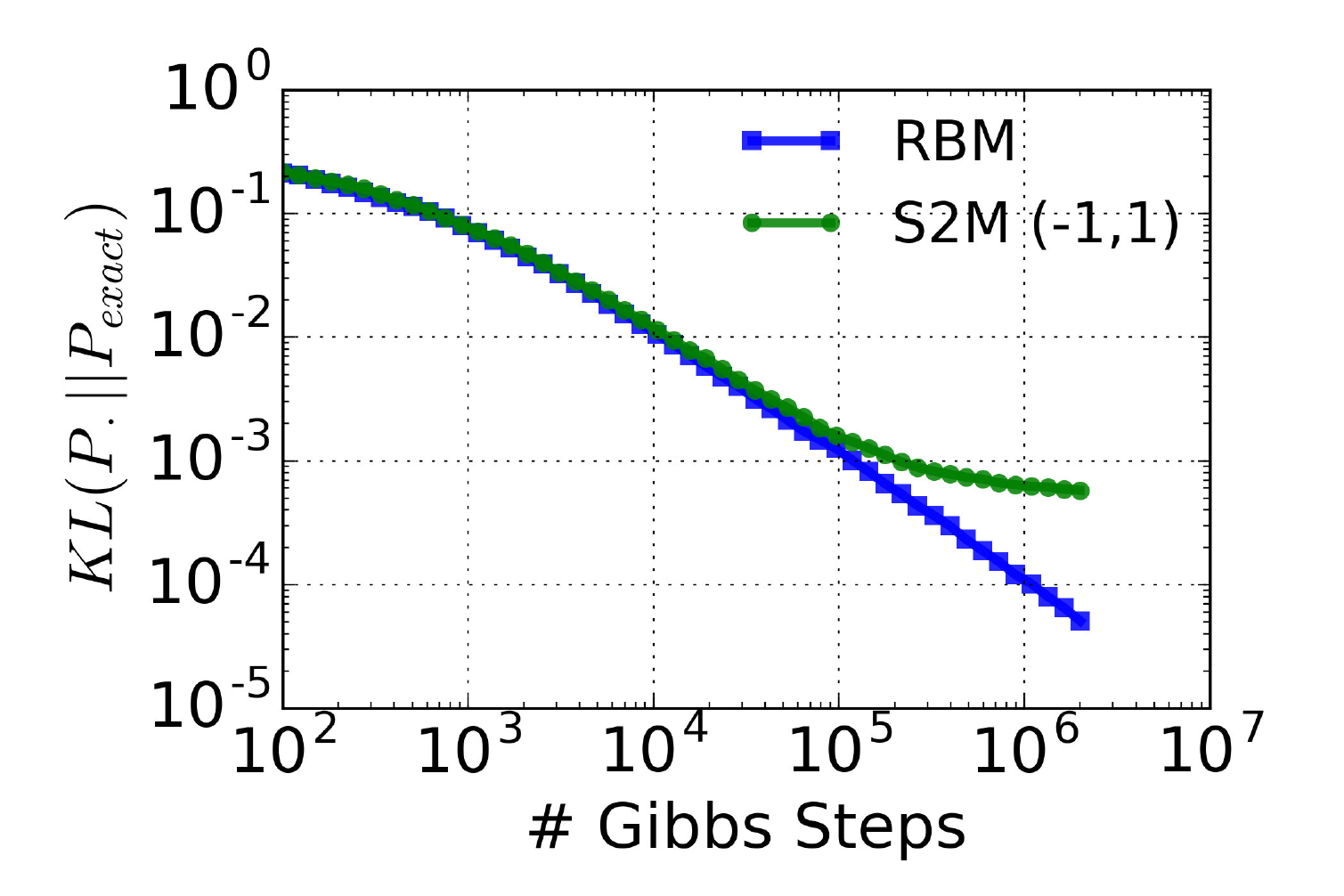}
    \caption{KL-divergence between a restricted Boltzmann distribution ($P_{exact}$, 5 hidden units, 5 visible units) and the distribution sampled by a \ac{dSSM} with $r_{\text{on}}=1, r_{\text{off}}=-1$ and matched parameters. As a reference, the KL-divergence using an \ac{RBM} is shown (blue curve). The saturation of the green curve is caused by the inexact activation function of the \ac{dSSM} unit (error function instead of the logistic function). }
    \label{fig:kl_divergence_bssm}
\end{figure}

In the general case, the last property described above suggest that the \ac{dSSM} cannot be expressed as an energy-based model (as in the case for Boltzmann machines), and therefore we cannot easily derive the joint distribution $P(\mathbf{z[t]})$.
Some insight can be gained in the particular case where $r_{\text{on}}=-r_{\text{off}}$.
In this case, the denominator in \refeq{eq:ensemble_hopfield} simplifies such that:
\begin{eqnarray}
    P(z_i[t+1] = 1| \mathbf{z}[t]) &=& P(u_i [t+1] \ge 0| \mathbf{z}[t])\\
    &=& \frac12 \left(1 + \mathrm{erf}\left(  \beta    \frac{\sum_j w_{ij} z_j[t] + \frac{b_i}{p}}{r_{\text{on}}\sqrt{\sum_j w_{ij}^2}} \right)\right).
\end{eqnarray}
In other words: the standard deviation of the inputs $\sigma_i^2$ does not depend on the neural states.
With the additional constraint that $\sum_{i} w_{ij}^2 = \sum_{j} w_{ij}^2$, the connectivity matrix becomes symmetric.
In this case, the resulting system is \emph{almost} a Boltzmann machine, with the only exception that the neural activation function is an error function instead of the logistic function.
Generally speaking, the error function is reasonably close to the logistic function after a rescaling of the argument.
Therefore the parameters of a Boltzmann machine can be approximately mapped on the \ac{dSSM} with $r_{\text{on}}=-r_{\text{off}}$. 
To test the quality of this approximation, we compute the Kullback-Leibler (KL) divergence between an exact restricted Boltzmann distribution ($P_{exact}$) and the distribution sampled by a \ac{dSSM} with $r_{\text{on}}=1, r_{\text{off}}=-1$ \reffig{fig:kl_divergence_bssm}. This computation is repeated in the case of the distribution obtained with Gibbs sampling in the \ac{RBM}.
In order to avoid zero probabilities, we added 1 to the number of occurrences of each state.
As expected, the KL-divergence between the \ac{dSSM} and the exact distribution reaches a plateau due to the non-logistic activation function.
However, the results show that in networks of this size the distribution sampled by the \ac{dSSM} has the same KL-divergence as the RBM obtained after $10^5$ iterations of Gibbs sampling, which is more than the typical number of iterations used for many tasks \cite{Hinton_etal06_fast-lear}.
We premise that the \acp{dSSM} with all or none activation values ($r_{\text{on}}\neq r_{\text{off}}=0$) behave in a manner that is sufficiently similar, so that learning in the \ac{dSSM} with \ac{CD} is approximately valid. 
In the Results section, we test this premise on the MNIST hand-written digit machine learning task.

\subsubsection*{Learning Rule for RBM and \acp{dSSM}}
The training of \acp{RBM} proceeds in two phases.
During the first ``data'' phase, the states of the visible units are clamped to a given vector of the training set, then the states of the hidden units are sampled.
In a second ``reconstruction'' phase, the network is allowed to run freely.
Using the statistics collected during sampling, the weights are updated in a way that they maximize the likelihood of the data.
Collecting equilibrium statistics over the data distribution in the reconstruction phase is often computationally prohibitive.
The \ac{CD} algorithm has been proposed to mitigate this \cite{Hinton02_trai-prod,Hinton_Salakhutdinov06_redu-dime}: the reconstruction of the visible units' activity is achieved by sampling them conditioned on the values of the hidden units.
This procedure can be repeated $k$ times (the rule is then called \ac{CD}$_k$), but relatively good convergence is obtained for the equilibrium distribution even for one iteration.
The \ac{CD} learning rule is summarized as follows:
\begin{equation}\label{eq:cd_rule}
    \Delta w_{ij} = \epsilon( \langle v_i h_j \rangle_{\text{data}} - \langle v_i h_j \rangle_{\text{recon}}),
\end{equation}
where $v_i$ and $h_j$ are the activities in the visible and hidden layers, respectively.
The notations $\langle\cdot\rangle_{data}$ and $\langle\cdot\rangle_{recon}$ refer to expectations computed during the ``data'' phases and the ``reconstruction'' sample phases, respectively.
The learning of the biases follows a similar rule.
The learning rule for the \ac{dSSM} was identical to that of the \ac{RBM}. 

\subsection{Spiking Synaptic Sampling Machines}
\subsubsection*{Neuron and Synapse Model}
Except for the noise model, the neuron and synapse models used in this work are identical to those described in \cite{Neftci_etal14_even-cont} and are summarized here for convenience.
The neuron's membrane potential below firing threshold $\theta$ is governed by the following differential equation:
\begin{equation}\label{eq:clif}
        C \frac{\mathrm{d}}{\mathrm{d}t} u_i = - g_L u_i + I_i(t),\quad u_i(t)\in(-\infty,\theta),
\end{equation}

where $C$ is a membrane capacitance, $u_i$ is the membrane potential of neuron $i$, $g_L$ is a leak conductance, $I_i(t)$ is the time-varying input current and $\theta$ is the neuron's firing threshold. 
When the membrane potential reaches $\theta$, an action potential is elicited.
After a spike is generated, the membrane potential is clamped to the reset potential $u_{rst}$ for a refractory period $\tau_{r}$.

In the \ac{S3M}, the currents $I_i(t)$ depend on the layer the neuron is situated in (visible $v$ or hidden $h$). 
For a neuron $i$ in the visible layer $v$
\begin{equation}\label{eq:Iv}
    \begin{split}
    I_i(t)  = & I^d_i(t) + I^v_i(t),\\
    \tau_{syn} \frac{\mathrm{d}}{\mathrm{d} t} I^{v}_i(t)  = & -I^{v}_i + \sum_{j=1}^{N_h} \xi^p_{h,ji}(t) q_{ji} h_j(t)  + b_{v_i},
    \end{split}
\end{equation}
where $I^d_{i}(t)$ is a current representing the data (\emph{i.e.} the external input), $I^v(t)$ is the feedback from the hidden layer activity and the bias. The $q$'s are the respective synaptic weights, $\xi^p_{v,ij}$ is a Bernoulli process (\emph{i.e.} ``coin flips'') with probability $p$ implementing the stochasticity at the synapse, and $b_{v_i}$ are constant currents representing the biases of the visible neurons.
Spike trains are represented by a sum of Dirac delta pulses centered on the respective spike times:
\begin{equation}\label{eq:delta_diracs}
h_j(t) = \sum_{k\in Sp_j^h} \delta(t-t_k), \quad v_i(t) = \sum_{k\in Sp_i^v} \delta(t-t_k)
\end{equation}
where $Sp^h_j$, $Sp^v_j$ are the sets the spike times of the hidden neuron $h_j$ and visible neurons $v_i$, respectively, and $\delta(t)=1$ if $t=0$ and $0$ otherwise.

For a neuron $j$ in the hidden layer $h$,
\begin{equation}\label{eq:Ih}
    \begin{split}
    I_j(t)  = & I^h_j(t),\\
    \quad  \tau_{syn} \frac{\mathrm{d}}{\mathrm{d} t} I^{h}_j(t)  = & -I^{h}_j + \sum_{i=1}^{N_v} \xi^p_{v,ij}(t) q_{ij} v_i(t) + b_{h_j},
    \end{split}
\end{equation}
where $I^h(t)$ is the feedback from the visible layer, and $v(t)$ are Poisson spike trains of the visible neurons defined in \refeq{eq:delta_diracs} and $b_{h_j}$ are the biases of the hidden neurons. 
The dynamics of $I^h$ and $I^v$ correspond to a first-order linear filter, so each incoming spike results in \acp{PSP} that rise and decay exponentially (\emph{i.e.} alpha-\ac{PSP}) \cite{Gerstner_Kistler02_spik-neur}.

\subsubsection*{Event-Driven \ac{CD} Synaptic Plasticity Rule}
Event-driven Contrastive Divergence is an online variation of \ac{CD} amenable for implementation in neuromorphic and brain-inspired hardware: by interpreting spikes as samples of a probability distribution, a possible neural mechanism for implementing \ac{CD} is to use \ac{STDP} synapses to carry out \ac{CD}-like updates.

The weight update in \ac{eCD} is a modulated, pair-based \ac{STDP} rule with learning rate $\epsilon_q$:
\begin{equation}\label{eq:ecd_rule}
    \frac{\mathrm{d}}{\mathrm{d}t} q_{ij} = \epsilon_q g(t)\,\mathrm{STDP}_{ij}(v_i(t),h_j(t))
\end{equation}
where $g(t) \in \mathbb{R}$ is a zero-mean global gating signal controlling the data vs. reconstruction phase, $q_{ij}$ is the weight of the synapse and $v_i(t)$ and $h_j(t)$ refer to the spike trains of neurons $v_i$ and $h_j$, defined as in \refeq{eq:delta_diracs}.
The same rule is applied to learn biases $\mathbf{b}_v$ and $\mathbf{b}_h$:
\begin{eqnarray}
    \frac{\mathrm{d}}{\mathrm{d}t} b_{v_{i}} &=& \epsilon_b \frac12 g(t)\,\mathrm{STDP}_{i}(v_i(t),v_i(t)),\\
    \frac{\mathrm{d}}{\mathrm{d}t} b_{h_{i}} &=& \epsilon_b \frac12 g(t)\,\mathrm{STDP}_{i}(h_i(t),h_i(t)),
\end{eqnarray}
where $\epsilon_b$ is the learning rate of the biases. 
The weight update is governed by a symmetric \ac{STDP} rule with temporal window $K(t)= K(-t), \forall t$:
\begin{equation}
    \begin{split}
        \mathrm{STDP}_{ij}(v_i(t),h_j(t)) = & v_i(t) A_{h_j}(t) + h_j(t) A_{v_i}(t),\\
A_{h_j}(t)     = & A \int^t_{-\infty} \mathrm{d}s K(t-s) h_j(s),\\
A_{v_i}(t)    = &   A \int^t_{-\infty} \mathrm{d}s K(s-t) v_i(s),\\
    \end{split}
\end{equation}
with $A>0$ defining the magnitude of the weight updates.

Although any symmetric learning window can be used, for simplicity, we used a nearest neighbor update rule where:
\[ K(t-s) =
\begin{dcases*}
	1 & \text{if $|t-s| < \tau_{STDP}$} \\
	0 & \text{otherwise}
\end{dcases*}
\]
In our implementation, updates are additive and weights can change polarity.
A global modulatory signal that is synchronized with the data clamping phase modulates the learning to implement \ac{CD}:
\begin{equation}\label{eq:canonical_g}
    g(t) =
      \begin{dcases*}
          1 & \text{if  $mod(t,2T)\in (\tau_{br}, T)$}\\
          -1 & \text{if  $mod(t,2T) \in (T+\tau_{br}, 2T)$}\\
          0 & \text{otherwise}.
      \end{dcases*},
\end{equation}
The signal $g(t)$ switches the behavior of the synapse from \ac{LTP} to \ac{LTD} (\emph{i.e.} Hebbian to Anti-Hebbian).
The temporal average of $g(t)$ vanishes to balance \ac{LTP} and \ac{LTD}. 
The modulation factor is zero during some time $\tau_{br}$, so that the network samples closer to its stationary distribution when the weights updates are carried out.
The time constant $\tau_{br}$ corresponds to a ``burn-in'' time of MCMC sampling and depends on the overall network dynamics.
\cite{Neftci_etal14_even-cont} showed that, when pre-synaptic neurons and post-synaptic neurons fire according to Poisson statistics, \ac{eCD} is equivalent to \acs{CD}.
The effective learning window is:
\[
    \epsilon = 2A \frac{T-\tau_{br}}{2T}.
\]
We note that the learning rate between the \ac{dSSM} and the spiking \ac{S3M} cannot be compared directly because there is no direct relationship between the synaptic weights $q$ and $w$.

In the \ac{S3M}, weight updates are carried out even if a spike is dropped at the synapse. This speeds up learning without adversely affecting the entire learning process because spikes dropped at the synapses are valid samples in the sense of the sampling process.
During the data phase, the visible units were driven with constant currents equal to the logit of the pixel intensity (bounded to the range $[10^{-5}, .98]$ in order to avoid infinitely large currents), plus a white noise process of low amplitude $\sigma$ to simulate sensor noise.

\subsubsection*{Synaptic Blank-out Noise}
Perhaps the simplest model of synaptic noise is the blank-out noise: for each spike-event, the synapse has a probability $p$ of evoking a \ac{PSP}, and a probability $(1-p)$ of evoking no response.
This is equivalent to modifying the input spike train to each synapse such that spikes are dropped (blanked-out) with probability $p$. 
In particular, for a Poisson spike train of rate $\nu$, the blank-out with probability $p$ gives a Poisson spike train with rate $p\nu$ \cite{Goldberg_etal01_prob-syna,Parzen99_stoc-proc}.
For a periodic (regular) spike train, stochastic synapses add stochasticity to the system.
The coefficient of variation of the \ac{ISI} becomes $\sqrt{1-p}$.
The periodic spike train thus tends to a Poisson spike train when $p\rightarrow 0$, with the caveat that events occur at integer multiples of the original \ac{ISI}.
Intuitively, the neural network cycles through deterministic neural integration and stochastic synapses.
\cite{Moreno-Bote14_pois-spik} found that the recurring process of adding spikes through neuronal integration and removing them through stochastic synapses causes the spike statistics to remain Poisson-like (constant Fano Factor) over a large dynamical range.
Synaptic stochasticity is thus a biologically plausible source of stochasticity in spiking neural networks, and can be very simply implemented in software and hardware \cite{Goldberg_etal01_prob-syna}.

\subsubsection*{Spiking Neural Networks with Poisson Input and Blank-out Synapses}\label{sec:methods_ssm_math}
This section describes the neural activation function of leaky \ac{IF} neurons.
 The collective dynamics of spiking neural circuits driven by Poisson spike trains is often studied in the diffusion approximation \cite{Wang99_syna-basi,Brunel_Hakim99_fast-glob,Brunel00_dyna-spar,Fusi_Mattia99_coll-beha,Renart_etal03_comp-neur,Deco_etal08_dyna-brai,Tuckwell05_intr-to}.
   In this approximation, the firing rates of individual neurons are replaced by a common time-dependent population activity variable with the same mean and two-point correlation function as the original variables, corresponding here to a Gaussian process.
 The approximation is true when the following assumptions are verified:
   1) the charge delivered by each spike to the post-synaptic neuron is small compared to the charge necessary to generate an action potential,
   2) there is a large number of afferent inputs to each neuron,
   3) the spike times are uncorrelated.
 In the diffusion approximation, only the first two moments of the synaptic current are retained.
 The currents to the neuron, $I(t)$, can then be decomposed as:
 \begin{equation}\label{eq:diff_approx_eq}
     I(t) = \mu + \sigma \eta(t),
 \end{equation}
 where $\mu = \langle I(t) \rangle = p \sum_j q_j \nu_j + b$ and $\sigma^2 = p \sum_j q_j^2 \nu_j$, $\nu_j$ is the firing rate of pre-synaptic neuron $j$, and  $\eta(t)$ is the white noise process. 
 Note that a Poisson spike train of mean rate $\nu$ with spikes removed with probability $p$ is a Poisson spike train with parameter $p \nu$.
 As a result, blank-out synapses have the effect of scaling the mean and the variance by $p$.

 In this case, the neuron's membrane potential dynamics is an Ornstein-Uhlenbeck process \cite{Gardiner12_hand-stoc}.
 For simplicity of presentation, the reset voltage was set to zero ($u_{rst}=0$).
 We consider the case where the synaptic time constant dominates the membrane time constant. 
 In other words, the membrane potential closely follows the dynamics of the synaptic currents and the effect of the firing threshold and the reset in the distribution of the membrane potential can be neglected.
 This problem was studied in great detail with first order approximations in $\tau_m/\tau_{syn}$ \cite{Petrovici_etal13_stoc-infe}.
 For comparison with the \ac{dSSM}, we focus here on the case $\tau_m=0$. 
 In this case, the stationary distribution is a Gaussian distribution:
 \[
     u \sim N(\frac{\mu}{g_L},\frac{\sigma^2}{2 g_L^2 \tau_{syn}}).
 \]
 Neurons such that $p(u>0|\mathbf{\nu})$ fire at their maximum rate of $\frac1{\tau_{ref}}$.
 Following similar steps as in the \ac{dSSM}, the firing rate of a neuron $i$ becomes:
 \begin{equation}\label{eq:tf_original}
     \begin{split}
     \nu_i & = \frac1{\tau_{ref}}\frac12\left(1+\operatorname{erf}\left( \frac{\mu_i}{\sigma_i}\sqrt{\tau_{syn}} \right)\right),\\
     & = \frac1{\tau_{ref}}\frac12\left(1+\operatorname{erf}\left( \sqrt{p} \frac{\sum_j q_{ij} \nu_j + \frac{b_i}{p}}{\sqrt{\sum_j q_{ij}^2 \nu_j}}\sqrt{\tau_{syn}} \right)\right).
     \end{split}
 \end{equation}

 \refeq{eq:tf_original} clarifies how the noise amplitude $\sigma$ affects the neural activation $\nu$, and thus allows a quantitative comparison with the \ac{dSSM}.
 Similarly to \refeq{eq:ensemble_hopfield}, the input is effectively normalized by the variability in the inputs (where on/off values are replaced by firing rates).
 These strong similarities suggest that the \ac{dSSM} and the spiking neural network are very similar.
 Using computer simulations of the MNIST learning task, in the Results section we show that the two networks indeed perform similarly.

\subsection{Training Protocol and Network Structure for MNIST}

\begin{figure}
   \begin{center}
   \includegraphics[width=0.35\textwidth]{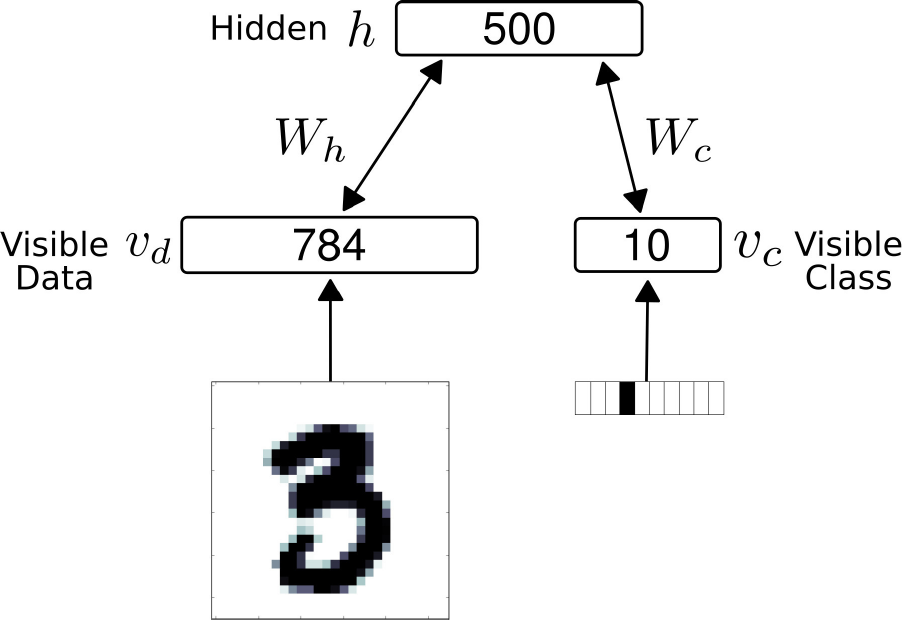}
   \end{center}
   \caption{\label{fig:network_architecture} The network architecture consists of a visible and a hidden layer.
   The visible layer is partitioned into $784$ sensory (data) neurons ($\mathbf{v_d}$) and $10$ class label neurons ($\mathbf{v_c}$). During data presentation, the activities in the visible layer are clamped to the data, consisting of a digit and its label (one-hot coded). In all models, the weight matrix between the visible layer and the hidden layer is symmetric.}
\end{figure}

\paragraph{\ac{RBM} and the \ac{dSSM}}
The network consisted of $1294$ neuron, partitioned into $794$ visible neurons and $500$ hidden neurons (\reffig{fig:network_architecture}).
In the results, we used $r_{on}=1$ and $r_{off}=0$ for the \ac{dSSM} neurons for a closer comparison with spiking neurons and the \ac{S3M}. 
The visible neurons were partitioned into $784$ sensory (data) neurons, denoted $\mathbf{v_d}$ and $10$ class label neurons, denoted $\mathbf{v_c}$.
The \ac{dSSM} is similar to the \ac{RBM} in all manners, except that the units are threshold functions, and each weight is multiplied by independently drawn binomials ($p=.5$) at every step of the Gibbs sampling.

In both cases the training set was randomly partitioned into batches of equal size $N_{batch}$.
One epoch is defined as the presentation of the full MNIST dataset (60000 digits).
We used CD-1 to train the \ac{RBM} and the \ac{dSSM}.

Classification is performed by choosing the most likely label given the input, under the learned model.
To estimate the discrimination performance, we sampled the distribution of class units $P(class | digit)$ using multiple Gibbs Sampling chains (50 parallel chains, 2 steps). 
We take the classification result to be the label associated to the most active class unit averaged across all chains.

For testing the discrimination performance of an energy-based model such as the \ac{RBM}, it is common to compute the free-energy $F(\mathbf{v_c})$ of the class units \cite{Haykin99_neur-netw}, defined as:
\begin{equation}
    \exp(-F(\mathbf{v_c})) = \sum_{\mathbf{v_d},\mathbf{h}} \exp(-E(\mathbf{v_d},\mathbf{v_c},\mathbf{h})),
\end{equation}
and selecting $\mathbf{v_c}$ such that the free-energy is minimized.
The free-energy computes the probabilities of a given neuron to be active, using the joint distribution of the \ac{RBM}.
Classification using free energy is inapplicable to \acp{dSSM} because it cannot be expressed in terms of an energy-based model (See Methods).
Therefore, throughout the article, free energy-based classification was used only for the \ac{RBM}.

The learning rate in the RBM and the \ac{dSSM} was linearly reduced during the training, reaching $0$ at the end of the learning.
Both \ac{RBM} and \ac{dSSM} were implemented on a GPU using the Theano package \cite{Bergstra_etal10_thea-cpu}.

\paragraph{Spiking \ac{dSSM}}
The \ac{S3M} network structure is identical to above.
Similarly to \cite{Neftci_etal14_even-cont}, for a given digit, each neuron in layer $\mathbf{v}$ was injected with currents transformed according to a logit function $I^d_i \propto \log \left(\frac{s_i}{1-s_i \tau_{r}}\right)$, where $s_i$ is the value of pixel $i$.
To avoid saturation of the neurons using the logit function, the pixel values of the digits were bounded to the range [$10^{-5},.98$]

Training followed the \ac{eCD} rule described in \refeq{eq:ecd_rule}. 
The learning rate was decayed linearly during the training, reaching $0$ at the end of the learning.
Similarly to the \ac{dSSM}, the discrimination performance is evaluated by sampling the activities of the class units for every digit in the test dataset.  
In the \ac{S3M}, this is equivalent to identifying the neuron that has the highest firing rate.
The spiking neural network was implemented in the spike-based neural simulator Auryn, optimized for recurrent spiking neural networks with synaptic plasticity \cite{Zenke_Gerstner14_limi-to}.
Connections in the \ac{SSM} are symmetric, but due to the constraints in the parallel simulator, the connections were implemented using two separate synapses (one in each direction), and periodically symmetrized to maintain symmetry (every $\unit[1000]{s}$ of simulated time).
A complete list of parameters used in our simulator is provided in the appendix \reftab{tab:parameters}.

\section{Results}

We demonstrate two different implementations of synaptic sampling machines, one spiking and one non-spiking.
The non-spiking \ac{dSSM} is a variant of the original \ac{RBM}, used to provide insight into the role of stochastic synapses in neural networks, and to justify \acp{S3M}. 
The \ac{S3M} consisted of a network of deterministic \ac{IF} spiking neurons, connected through stochastic (blank-out) synapses (\reffig{fig:srbm}).
In the Methods section, we show that \acp{dSSM} with activation values $(-1,+1)$ are similar to Boltzmann machines, except that the activation function of the neuron is an error function instead of the logistic function, and where the input is invariant to rescaling.
We assumed that the \ac{dSSM} with activation levels (0,1) and the \ac{S3M} are sufficiently similar to the Boltzmann distribution, such that CD and eCD as originally described in \cite{Neftci_etal14_even-cont} are approximately valid.
We test this assumption through computer simulations of the \acp{SSM} in an MNIST hand-written digit learning task.
Independently, M\"uller explored the idea of unreliable connections in Hopfield units, reaching similar conclusion on the role of the blank-out probabilities in the Boltzmann temperature, as well as such networks being good approximations of restricted Boltzmann machines \cite{Muller15_algo-mass}.
\subsection{Unsupervised Learning of MNIST Handwritten Digits in \aclp{SSM}}
Similarly to \acp{RBM}, \acp{SSM} can be trained as generative models.
Generative models have the advantage that they can act simultaneously as classifiers, content-addressable memories, and carry out probabilistic inference.
We demonstrate these features in a MNIST hand-written digit task \cite{LeCun_etal98_grad-lear}, using networks consisting of two layers, one visible  and one hidden (\reffig{fig:network_architecture}).

\begin{figure}
    \centering
    \includegraphics[width=0.4\textwidth]{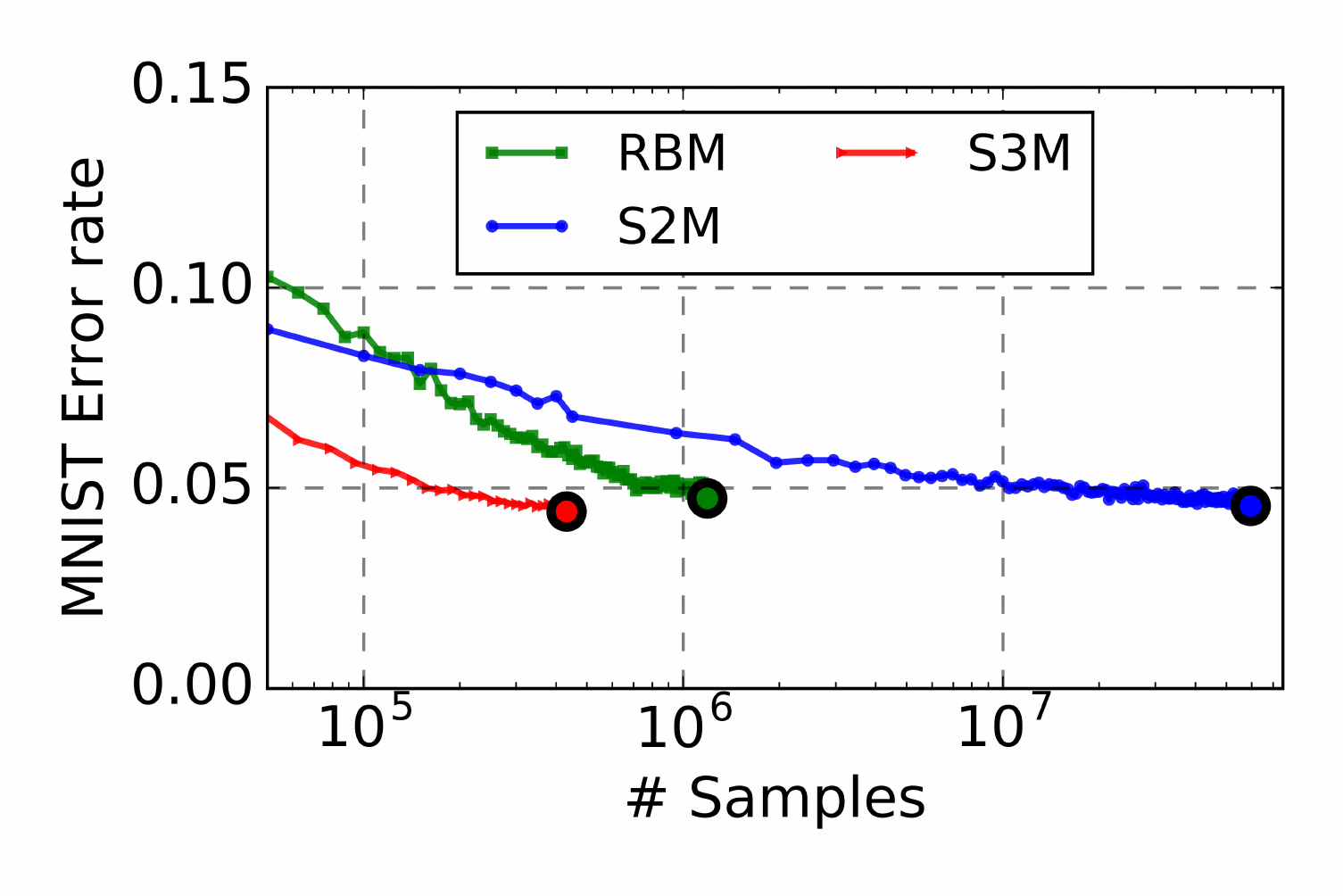}
    \caption{\label{fig:fig_convergence} \emph{The \ac{SSM} outperforms its \ac{RBM} counterpart at the MNIST task}. The \ac{RBM}, the \ac{dSSM} and the continuous-time \ac{S3M} were trained to learn a generative model of MNIST handwritten digits. The \ac{dSSM} model is identical to the \ac{RBM} except that it consists of threshold units with stochastic blank-out synapses with probability $50\%$. The recognition performance is assessed on the testing dataset which was not used during training (10000 digits). Error rate in the \ac{RBM} starts increasing after reaching peak performance, mainly due to decreased ergodicity of the Markov chains and overfitting. Learning in the \ac{dSSM} is slower than in the RBM, as reported with other models using DropConnect \protect\cite{Wan_etal13_regu-neur} but it is effective in preventing overfitting. At the end of the training, the recognition performance of the \acp{S3M} (S3M) averaged over 8 runs with different seeds reached 4.6\% error rate. Due to the computational load of running the spike-based simulations on the digital computer, the \ac{S3M} was halted earlier than the \ac{RBM} and the \ac{dSSM}. In spite of the smaller number of digit presentations, the \ac{S3M} outperformed the \ac{RBM} and the \ac{dSSM}. This is partly because weight updates in the \ac{S3M} are undertaken during each digit presentation, rather than after each minibatch. The curves are plotted up to the point where the best performance is reached.}
\end{figure}

Learning results are summarized in \reftab{tab:results}.
The \ac{S3M} appears to slightly outperform the \ac{dSSM}, although a direct comparison between the two is not possible because the sampling mechanism and the batch sizes are different.
We find that \acp{S3M} attain classification error rates that slightly outperform the machine learning algorithm (error rates: \ac{S3M} 4.4\% \emph{vs.}~RBM 5\%), even after much fewer repetitions of the dataset (\reffig{fig:fig_convergence}).
To date, this is the best performing spike-based unsupervised learner on MNIST.
The spiking network implementing the \ac{S3M} is many times smaller than the best performing spike-based unsupervised learner to date (1294 neurons 800k synapses \emph{vs.} 7184 neurons, 5M synapses) \cite{Diehl_Cook15_unsu-lear}.
For comparisons with other recent techniques for unsupervised learning in spiking neural networks, we refer the reader to \cite{Diehl_Cook15_unsu-lear}, which provides a recent survey of the MNIST learning task with spiking neural networks.

\begin{table*}
\begin{center}
    {\scriptsize
  \begin{tabular}{l  c  c }
    Model, $n_H$ = 500, 60k digits training set                          & MNIST Error \\
    \hline
    \ac{S3M} + Event-driven \ac{CD} (this work)                               &\textbf{4.4}\%\\
    \ac{S3M} + Event-driven \ac{CD} + 74\% connection pruning + relearning (this work)                               &\textbf{5.0}\%\\
    \ac{S3M} + Event-driven \ac{CD} + 4 bit synaptic weights (this work)   &5.2\%\\
    \ac{S3M} + Event-driven \ac{CD} + 2 bit synaptic weights (this work)   &7.8\%\\
    \ac{dSSM} + Standard \ac{CD} (this work)                     &\textbf{4.5}\%\\
    Gibbs Sampling + Standard \ac{CD}                                            &4.7\%\\
    Neural Sampling + Event-driven \ac{CD} \cite{Neftci_etal14_even-cont}                     &8.1\%\\
    Neural Sampling + Event-driven \ac{CD} \cite{Neftci_etal14_even-cont} (5 bits post-learning rounding)                  & 10.6\%\\
    \hline
  \end{tabular}
  }
\end{center}
\caption{\label{tab:results}}
\end{table*}

\begin{figure}
    \centering
    \includegraphics[width=.35\textwidth]{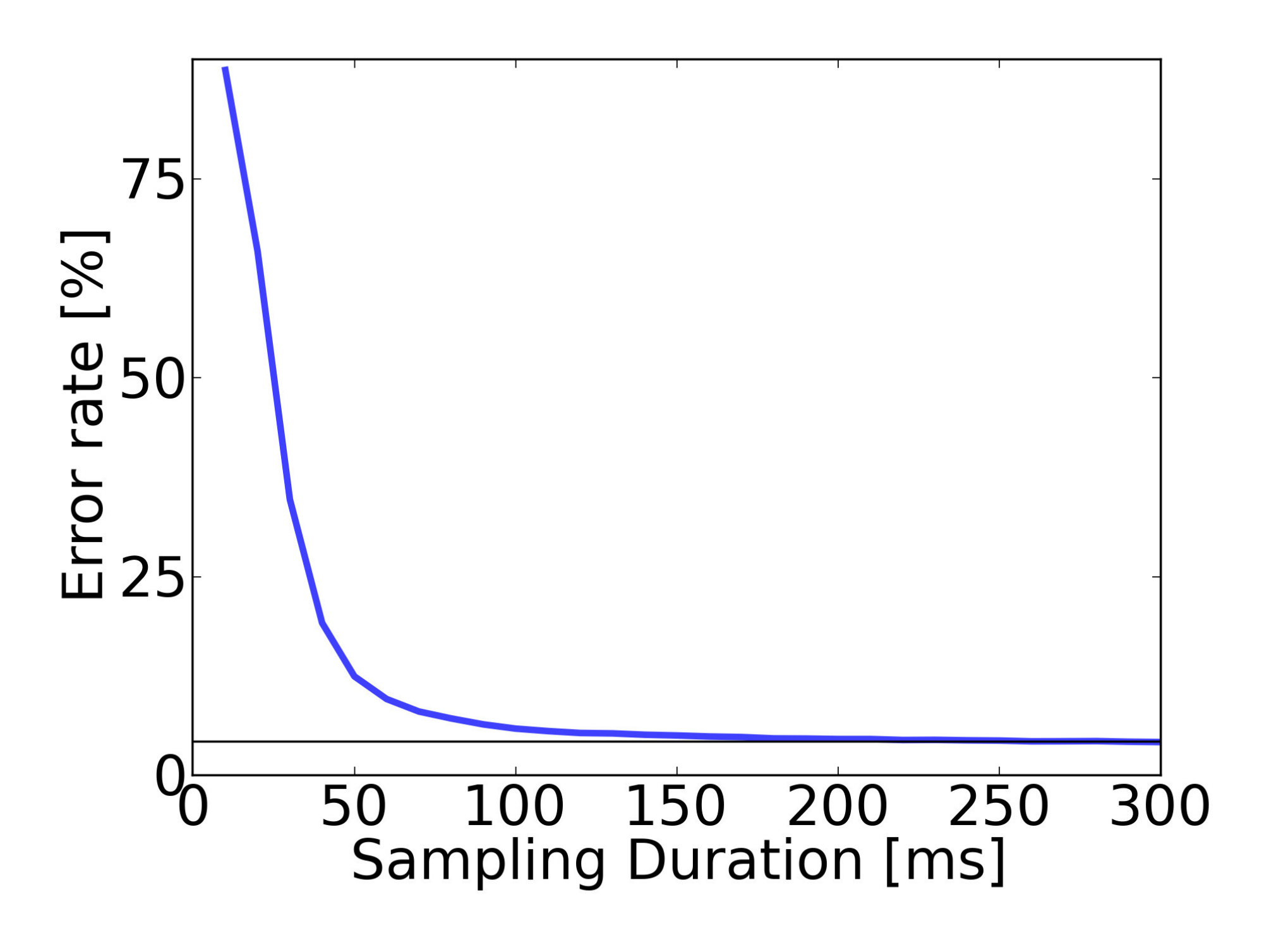}
    \caption{\label{fig:performance} \emph{Accuracy evaluation in the \ac{S3M}}. To test recognition, for each digit in the test dataset (10000 digits), class neurons in the \acp{SSM} are sampled for up to $\unit[300]{ms}$. The classification is read out by identifying the group of class label neurons that had the highest activity and averaging over all digits of the test set. The error rate after $\unit[50]{ms}$ of sampling was above $13\%$ and after $\unit[250]{ms}$ the error rates typically reached their minimum for this trained network ($4.2\%$, horizontal bar). }
\end{figure}

We tested the speed of digit classification under the trained \ac{SSM}.
We computed the prediction after sampling for a fixed time window that we varied from $0$ms to $300$ms (\reffig{fig:performance}).
Results show that the trained network required approximately $250$ms to reach the lowest error rate of $4.2$\%, and that only $13$\% of the predictions were incorrect after $50$ms. 
%

\begin{figure*}
    \centering
    \includegraphics[width=.8\textwidth]{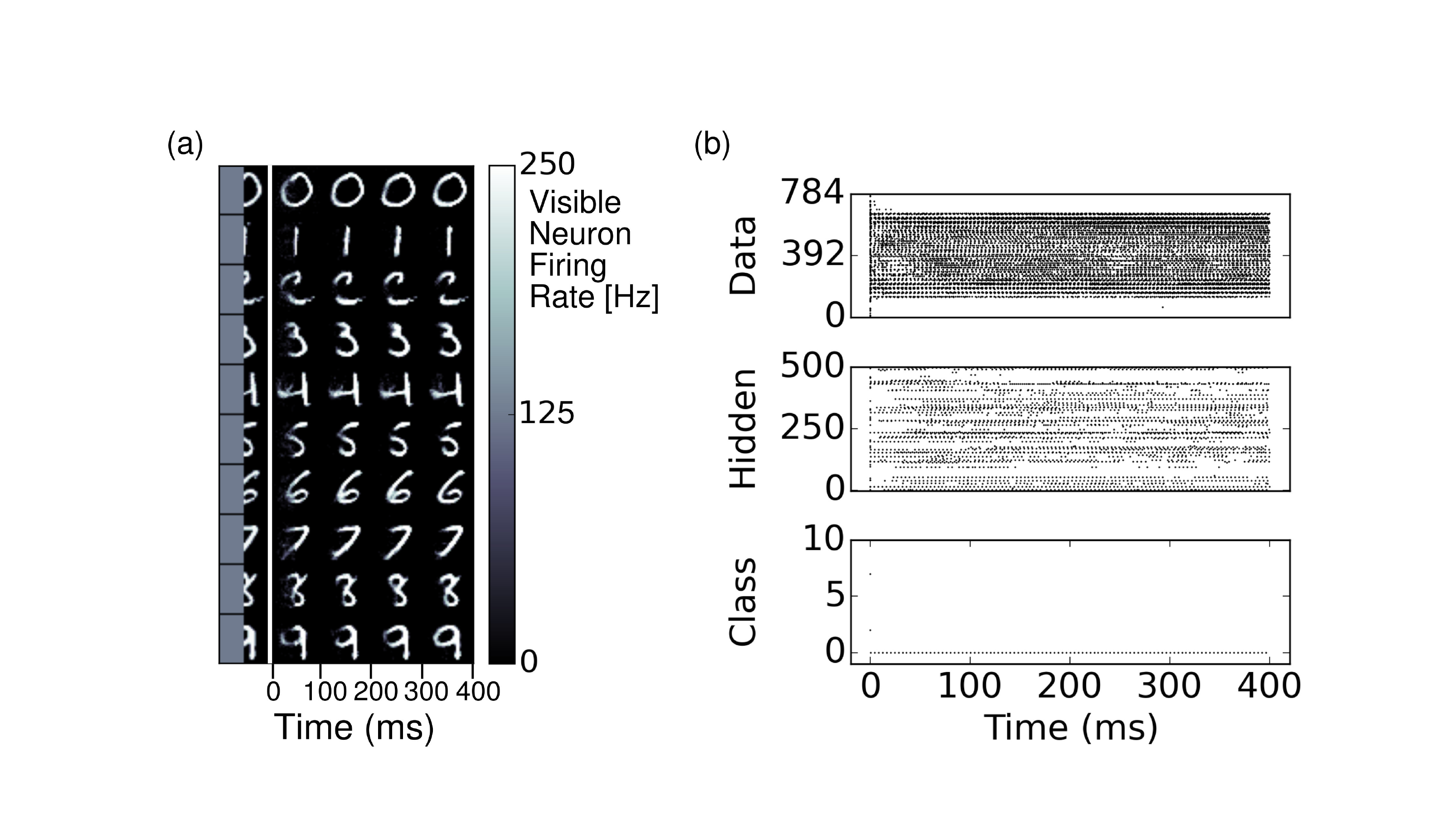}
    \caption{\label{fig:completion_compose} Pattern completion in the \ac{S3M}. (a) The right half of a digit and its label is presented to the visible layer of the \ac{SSM}. The ensuing activity in the visible layer is partitioned in $\unit[100]{ms}$ bins and the average activity during each bin is presented in the color plots to the right. The network completes the left half of the digit. (b) Raster plot of the pattern completion for the first row of panel (a).}
\end{figure*}

Similarly to \acp{RBM}, the \ac{SSM} learns a generative model of the MNIST dataset. 
This generative model allows to generate digits and reconstruct them when a part of the digit has been occluded.
We demonstrate this feature in a pattern completion experiment where the right half of each digit was presented to the network, and the visible neurons associated to the left half of the digit were sampled (\reffig{fig:completion_compose}).
In most cases, the \ac{SSM} correctly completed the left half of the digit.
\reffig{fig:completion_compose} also illustrates the dynamics of the completion, which appears to reach a stationary state after about $200$ms. 
Overall, these results show that the \ac{SSM} can achieve similar tasks as in \acp{RBM}, at a similar or better recognition performance while requiring fewer dataset iterations during learning.
\subsection{Representations Learned with Synaptic Sampling Machines are Sparse}
In deep belief networks, discriminative performance can improve when using binary features that are only rarely active \cite{Nair_Hinton09_3d-obje}.
Furthermore, in hardware sparser representations result in lower communication overhead, and therefore lower power consumption.

\begin{figure}
    \centering
    \includegraphics[width=.4\textwidth]{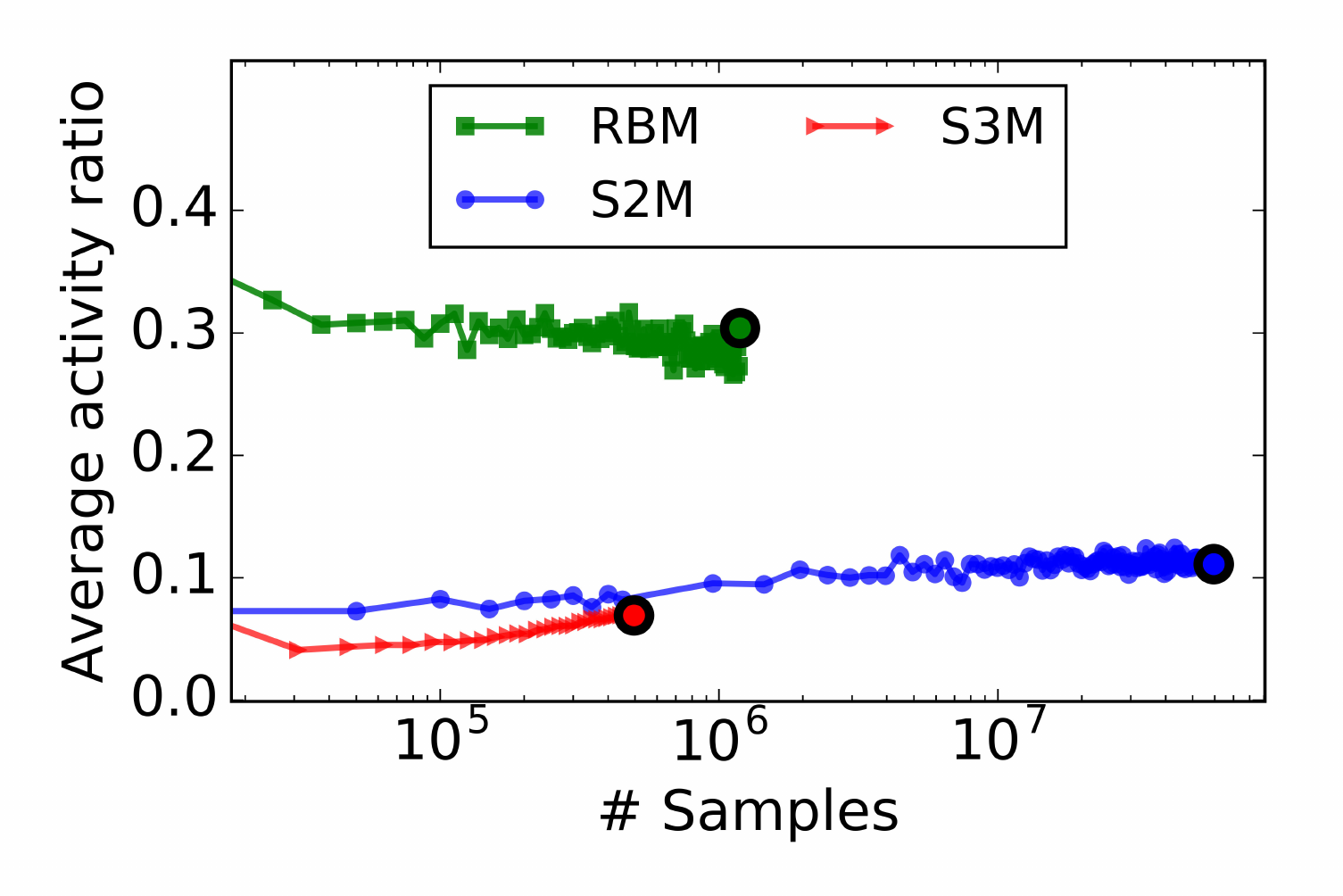}
    \caption{\label{fig:sparsity} \emph{For a similar parametrization, \acp{SSM} learn sparser representations than \acp{RBM}}.  The curves plot the average number of active units during learning in the hidden layer, computed over 50 digits for the \ac{dSSM} and the RBM, and 1000 digits for the \ac{S3M}. In the case of the \ac{S3M} (S3M), the average activity ratio is the average firing rate of the neurons in the network divided by the maximum firing rate $t_{ref}^{-1}$. The colored dot indicates the point where the peak of the recognition accuracy was reached. Nearly three times fewer units were active in the \ac{SSM} than in the \ac{RBM} at the designated points, and even fewer in the case of the \ac{S3M}.}
\end{figure}

To quantify the degree of sparsity in the \ac{RBM} and the \ac{SSM}, we computed the average fraction of active neurons.
For a similar parametrization, the average activity ratio of \acp{SSM} is much lower than that of \acp{RBM} (\reffig{fig:sparsity}).
RBMs can be trained to be sparse by added sparsity constraints during learning \cite{Lee_etal08_spar-deep}, but how such constraints can map to spiking neurons and synaptic plasticity rules is not straightforward.
In this regard, it is remarkable that \acp{SSM} are sparse without including any sparsity constraint in the model.
The difference in S2M and the S3M curves is likely to be caused by different parameters in the spiking and the non-spiking \ac{dSSM}, as these cannot be exactly matched.

\begin{figure*}
    \centering
    \includegraphics[width=.9\textwidth]{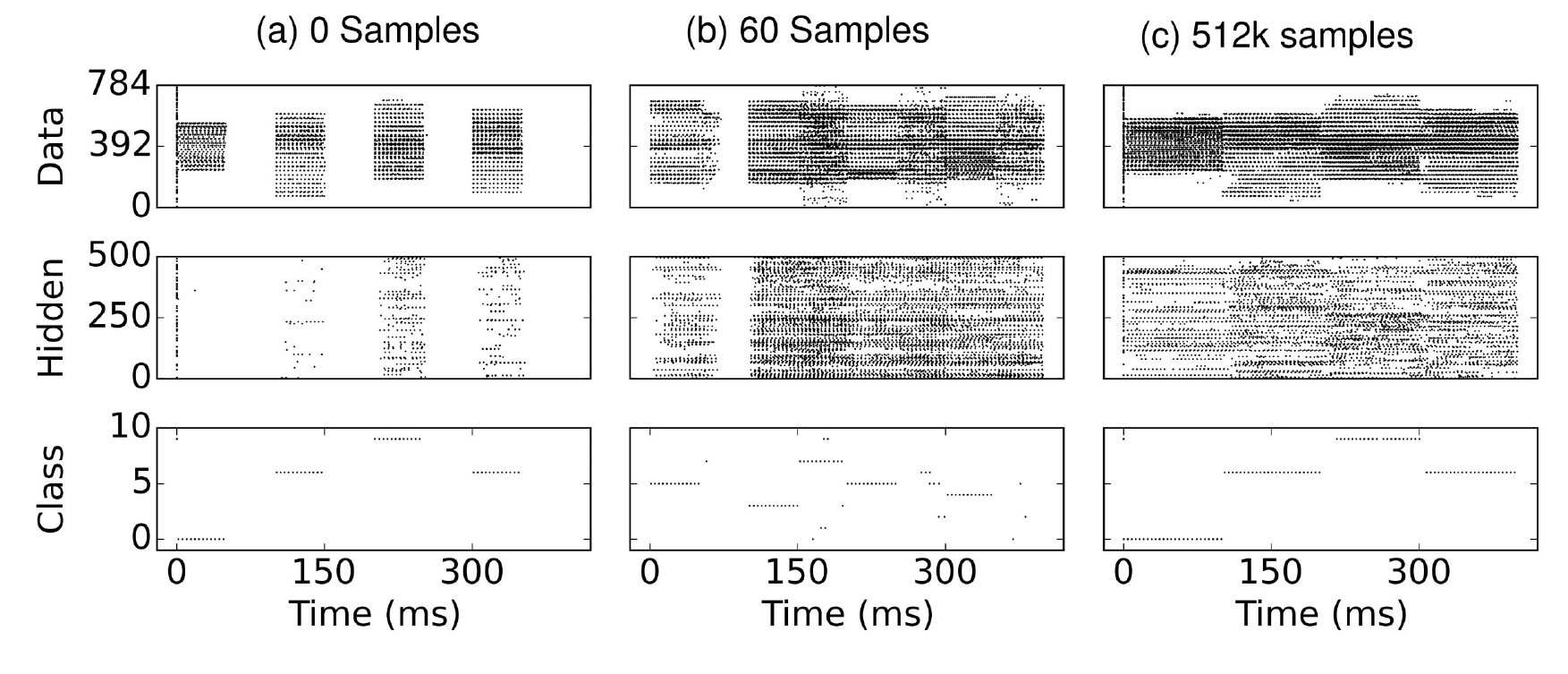}
    \caption{\label{fig:learning}(a-c) Spike rasters during \ac{eCD} learning. The data phase is $\unit[0-50]{ms}$ and the reconstruction phase is $\unit[50-100]{ms}$, repeated every $\unit[100]{ms}$ for a different digit of the training set. Due to the deterministic neural dynamics, a neuron receiving no input can never fire if its bias is below the rheobase (\emph{i.e.} the minimum amount of current necessary to make a neuron fire).
    Consequently, at the beginning of learning, the hidden neurons are completely silent in the reconstruction phases, and are gradually recruited during the data phase of the \ac{eCD} rule (\emph{e.g.} see \unit[50]{ms} in panel (b)).}
\end{figure*}

One can gain an intuition on the cause for sparsity in the \ac{SSM} by examining the network states during learning:
In the absence of additive noise, the input-output profile of leaky \ac{IF} neurons near the rheobase (minimal current required to cause the neuron to spike) is rectified-linear.
Consequently, without positive inputs and positive bias values, the spiking neuron cannot fire, and thus the pre-synaptic weights cannot potentiate.
By selecting bias values at or below zero, this feature causes neurons in the network to be progressively recruited, thereby promoting sparsity.
This is to be contrasted in the case of neurons with additive noise (such as white noise with constant amplitude), which can fire even if the inputs are well below rheobase.
In the \ac{SSM}, the progressive recruitment can be observed in \reffig{fig:learning}: early in the training, hidden neurons are completely silent during the reconstruction phase.
Because there is no external stimulus during the reconstruction phase, only neurons that were active during the data phase are active during the reconstruction phase.
After the presentation of 60 digits, the activity appears to ``grow'' from the data phase.
Note that similar arguments can be made in the case of \acp{dSSM} implemented with threshold neurons.

\subsection{Robustness of \acp{S3M} to Synapse Pruning and Weight Down-sampling}\label{sec:pruning}
Sparse networks that require using very few bits for storage can have a lower memory cost, along with a smaller communication and energy footprint.
To test the storage requirements for the \ac{S3M} and its robustness to pruning connections, we removed all synapses whose weights were below a given threshold.
We varied the threshold such that the ratio of remaining connections spanned the range $[0\%,100\%]$.

\begin{figure}
    \centering
    \includegraphics[width=.4\textwidth]{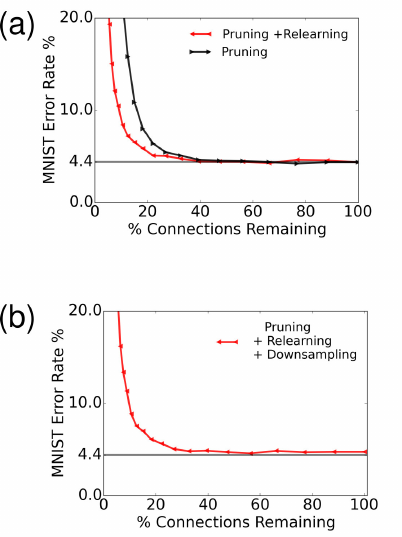}
    \caption{\label{fig:pruning-results} \emph{Pruning connections in the \ac{S3M} and re-learning.} (a) We test the \ac{S3M}'s robustness to synapse pruning by removing all connections whose weights were below a given threshold.
    We quantified the effects of pruning by testing the pruned \ac{S3M} over the entire MNIST test set (black curve).
    After pruning, we re-trained the network over 32 epochs (corresponding to 32k sample presentations), which recovered most of the loss due to pruning (red curve).
    (b) Reduced synaptic weight precision in the \ac{S3M}. After learning, pruning and re-learning, the synaptic weights were downsampled by truncating them to a single decimal point.
    There remained fewer than 128 unique weight values in the final matrix. The horizontal line is at 4.4\%.}
\end{figure}

The resulting recognition performance, plotted against the ratio of connections retained, is shown in \reffig{fig:pruning-results}a (black curve).
We then re-trained the \ac{S3M} over $32$ epochs and tested against $10,000$ images from the MNIST test dataset.
The re-learning substantially recovered the performance loss caused by the weight pruning as shown in \reffig{fig:pruning-results}a (red curve).
This result suggests that only a relatively small number of connections play a critical role in the underlying model for the recognition task.

\subsubsection*{Learning Low Precision Weight Synapses}
Dedicated memory for storing the synaptic connectivity table and the synaptic weights often occupies the largest area in neuromorphic devices \cite{Moradi_Indiveri13_even-neur,Merolla_etal14_mill-spik}.
Therefore, the ability to reduce the synaptic precision required for the operation of an algorithm can be very beneficial in hardware.
We quantify the effects of lowered precision in inference by downsampling the synaptic weights, while performing computation at full precision.
In the context of a hardware implementation, this results in lower memory costs since fewer bits can be used to store the same network.
To test the performance of the \ac{S3M} with lowered resolution, we truncated the weights to a single decimal point.
The resulting weights were restricted to less than 128 unique values (7 bits).
In \reffig{fig:pruning-results}b, we truncated the weights of the previously pruned network to 7 bits, and examined the results over the range of retained connections. The results of testing 10,000 samples of MNIST on this network are shown in \reffig{fig:pruning-results}b. The error rate was about 8\% with 12.5\% of the total synaptic weights retained and about 5\% with 49\% of the connections retained.

Another recently proposed technique for reducing the precision of the weights while minimizing impact on performance is the dual-copy rounding technique \cite{Stromatias_etal15_robu-spik}.
In the context of our sampling machine, the key idea is to sample using reduced precision weights, but learn with full precision weights.
Dual copy rounding was shown to outperform rounding of the weight after learning.

Training the \ac{S3M} with dual-copy rounding at 4-bit weights (16 different weight values) resulted in $5.2\%$ error rate at the MNIST task, and rounding at $2$ bits ($4$ weight values) increased this number to $7.8\%$.
Synaptic weight resolution of 4 bits is recognized as a sweet spot for hardware \cite{Merolla_etal14_mill-spik,Pfeil_etal12_4-bi-syna}.
Furthermore, it is a plausible synaptic weight precision for biological synapses:
recent analysis of synapses in the rat Hippocampus suggested that each synapse could store about 4 to 5 bits of information \cite{Bartol_etal15_hipp-spin}.

We note that the dual-copy approach is only beneficial at the inference stage (post-learning). 
During inference, the high-precision weights can be dropped and only the low-precision weight are maintained.
However, during learning it is necessary to maintain full precision weights.
Learning with low precision weights is possible using stochastic rounding techniques \cite{Muller_Indiveri15_roun-meth}.
We could not test stochastic rounding in spiking \acp{SSM} because the symmetry requirements in the spiking neural network connectivity prevent a direct, efficient implementation in multithreaded simulators (such as the used Auryn neural simulator).

\subsection{Synaptic Operations and Energy Efficiency in \acp{SSM}}
Power consumption in brain-inspired computers is often dominated by synaptic communication and plasticity.
Akin to the energy required per multiply accumulate operation (MAC) in digital computers, the energy required by each synaptic operation (SynOp) is one representative metric of the efficiency of brain-inspired computers \cite{Merolla_etal14_mill-spik}.
The reason is that every time one neuron spikes, a large number of synapses of other, possibly physically distant neurons are updated (in practice, hundreds to ten of thousands of synapses).

\begin{figure}
\centering
\includegraphics[width=0.5\textwidth]{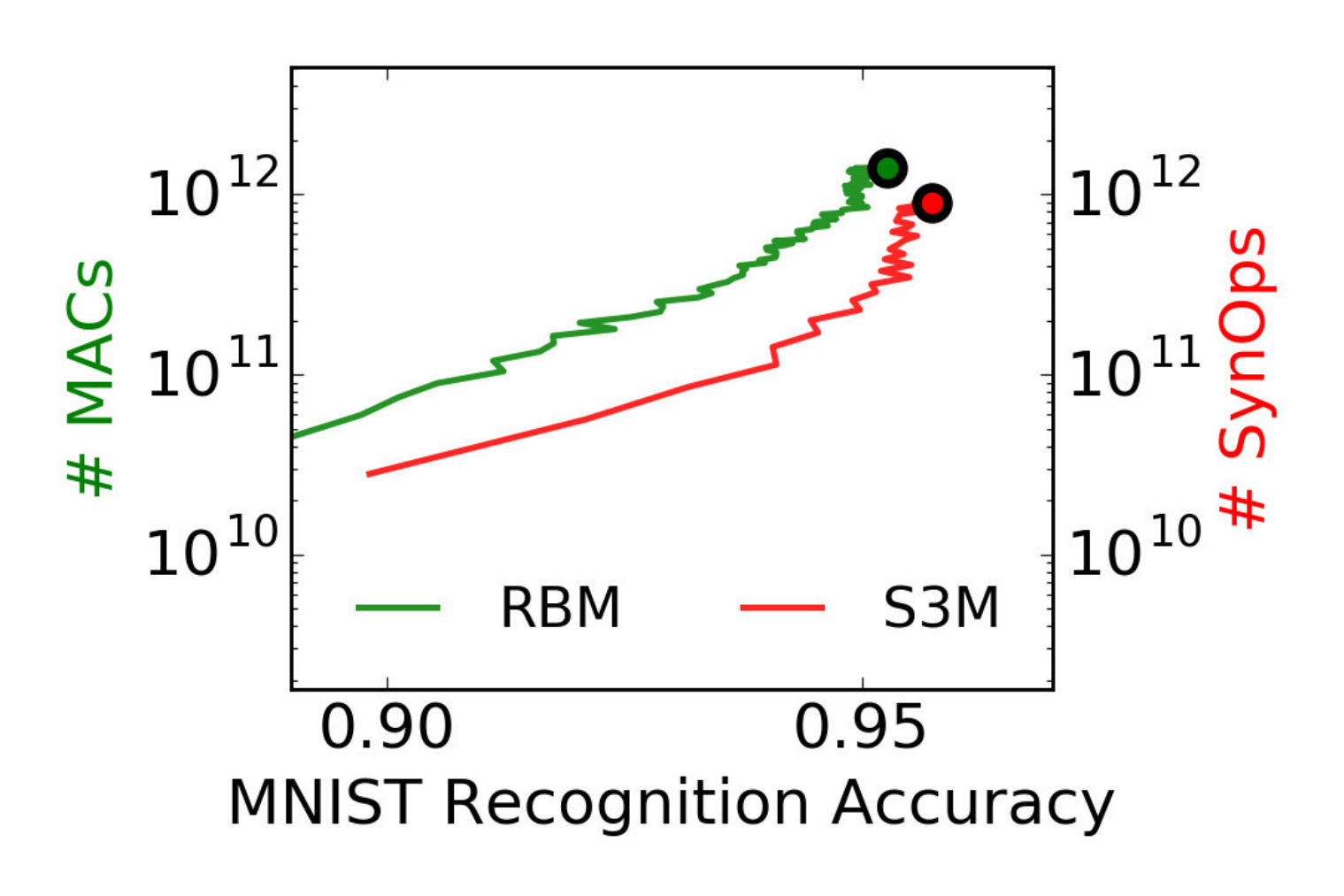}
  \caption{\label{fig:ann_snn_mnist_comparison} \emph{The \ac{S3M} achieves SynOp-MAC parity at the MNIST task}. The number of multiply-accumulate (MAC) operations required for sampling in the RBM during learning is compared to the number of synaptic operations (SynOps) in the \ac{S3M} during the MNIST learning task. At this task, the \ac{S3M} (S3M) requires fewer operations to reach the same accuracy during learning. Other necessary operations for the RBM, \emph{e.g.} additions, random number generations, logistic function calls and weight updates were not taken into account here, and would further favor the \ac{S3M}. One reason for the SynOp-MAC parity is that learning in the \ac{S3M} requires fewer repetitions of data samples to reach the same accuracy compared to the \ac{RBM}. Another reason is that only active connections in the \ac{S3M} incur a SynOp. SynOp-MAC parity between the \ac{S3M} and the \ac{RBM} is very promising for hardware implementations because a SynOp in dedicated hardware potentially consumes much less power than a MAC in a general purpose digital processor. Note that the non-spiking \ac{dSSM} is not competitive on this measure because it requires $N^2$ random number generations per Gibbs sampling step in addition to the same number of MACs as the RBM. The curves are plotted up to the point where the best performance is reached.}
\end{figure}

A SynOp potentially consumes much less energy than a MAC even on targeted GPGPUs, and improvements in energy per SynOP directly translate into energy efficiencies at the level of the task. 
However, these operations are not directly comparable because it is unclear how many SynOps provide the equivalent of a MAC operation at a given task.
To provide a reference to this comparison, we count the number of SynOps and approximate number of MACs necessary to achieve a target accuracy at the MNIST task for \acp{S3M} and RBMs, respectively  (\reffig{fig:ann_snn_mnist_comparison}).
Strikingly, the \ac{S3M} achieves SynOp-MAC parity at this task.
Given that the energy efficiency of a SynOp is potentially orders of magnitude lower than the MAC \cite{Park_etal14_65k--73-m,Merolla_etal14_mill-spik}, this result makes an extremely strong case for hardware implementations of \acp{S3M}.
The possible reasons for this parity are twofold: 1) weight updates are undertaken after presentation of every digit, which mean that fewer repetitions of the dataset are necessary to achieve the same performance.
2) only active connections incur a SynOp.
In the \ac{RBM}, the operations necessary for computing the logistic function, random numbers at the neuron, and the weight updates were not taken into account and would further favor the \ac{S3M}.
The operation necessary for the stochastic synapse in the \ac{S3M} (a Bernoulli trial) is likely to be minimal because the downstream synapse circuits do not need to be activated when the synapse fails to transmit.
Furthermore, the robustness of \acp{S3M} to the pruning of connections (described in \refsec{sec:pruning}) can further strongly reduce the number of SynOps after learning.

\section{Discussion}
The Boltzmann Machine stems from the idea of introducing noise into a Hopfield network \cite{Hinton_Sejnowski86_lear-rele}.
The noise prevents the state from falling into local minima of the energy, enabling it to learn more robust representations while being less prone to overfitting.
Following the same spirit, the \ac{SSM} introduces a stochastic component to Hopfield networks, but rather than the units themselves being noisy, the connections are noisy.

The stochastic synapse model we considered is blank-out noise, where the synaptic weight is multiplied by binary random variable.
This is to be contrasted with additive noise, where the stochastic term such as a white noise process is added to the membrane potential.
In artificial neural networks, multiplicative noise forces weights to become either sparse or invariant to rescaling \cite{Nalisnick_etal15_scal-mixt}.
When combined with rectified linear activation functions, multiplicative noise tends to increase sparsity \cite{Rudy_etal14_neur-netw}.
Multiplicative noise makes neural responses sparser: In the absence of additive noise, neurons have activation functions very close to being rectified linear \cite{Tuckwell05_intr-to}.
In the absence of a depolarizing input, the neuron cannot fire, and thus its synapses cannot potentiate.
Consequently, if the parameters are initiated properly, many neurons will remain silent after learning.

Event-driven \ac{CD} was first demonstrated in a spiking neural network that instantiated a \ac{MCMC} neural sampling of a target Boltzmann distribution \cite{Neftci_etal14_even-cont}.
The classification performance of the original model was limited by two properties.
First, every neuron was injected with a (additive) noisy current of very large amplitude.
Neurally speaking, this corresponded to a background Poisson spike train of $\unit[1000]{Hz}$, which considerably increased the network activity in the system.
Second, in spite of the large input noise, neurons fired periodically at large firing rates due to an absolute refractory period.
This periodic firing evoked strong spike-to-spike correlations (synchrony) that were detrimental to the learning and the sampling.
Consequently, the performance of \ac{eCD} in an MNIST task was significantly lower than when standard \ac{CD} was used (8.1\% error rate).

The performance of the \ac{S3M} in the MNIST hand-written digits task vastly improved accuracy metrics over our previous results while requiring a fraction of the number of synaptic operations.  
The reduction in the number of operations was possible because our previous neuron model required an extra background Poisson spike train for introducing noise, whereas the \ac{dSSM} generates noise through stochastic synapses.
The improvement in accuracy over our earlier results in \cite{Neftci_etal14_even-cont} stems from at least two reasons: 1) Spike-to-spike decorrelations caused by the synaptic noise better condition the plasticity rule by preventing pair-wise synchronization;
2) Regularization, which mitigates overfitting and the co-adaptation of the units.
In the machine learning community, blank-out noise is also known as DropConnect \cite{Wan_etal13_regu-neur}.
It was demonstrated to perform regularization, which helped achieve state-of-the-art results on several image recognition benchmarks, including the best result so far on MNIST without elastic distortions ($0.21\%$).
Note that the \ac{SSM} model did not include domain-specific knowledge, which suggests that its performance may generalize to other problems and datasets.

\subsection{Synaptic Unreliability in the Brain}
The probabilistic nature of synaptic quantal release is a well known phenomenon \cite{Katz66_nerv-musc}.
Unreliability is caused by the probabilistic release of neurotransmitters at the pre-synaptic terminals \cite{Allen_Stevens94_eval-caus}.
Detailed slice and \emph{in vivo} studies found that synaptic vesicle release in the brain can be extremely unreliable - typically 50\% transmission rate and possibly as low as 10\% in \emph{in vivo} - at a given synapse \cite{Branco_Staras09_prob-neur,Borst10_low-syna}.
Such synaptic unreliability can be a major source of noise in the brain \cite{Calvin_Stevens68_syna-nois,Faisal_etal08_nois-nerv,Yarom_Hounsgaard11_volt-fluc,Abbott_Regehr04_syna-comp}.
Assuming neurons maximize the ratio of information theoretic channel capacity to axonal transmission energy, synaptic failures can lower energy consumption without lowering the transmitted computational information of a neuron \cite{Levy_Baxter02_ener-neur}. 
Interestingly, an optimal synaptic failure rate can be computed given the energetic cost synaptic and somatic activations.
Another consequence of probabilistic synapses is that, via recurrent interactions in the networks, synaptic unreliability would be the cause of Poisson-like activity in the pre-synaptic input \cite{Moreno-Bote14_pois-spik}.

In the \ac{SSM}, the multiplicative effect of the blank-out noise is manifested in the pre-synaptic input by making its variance dependent on the synaptic weights and the network states.
Our theoretical analysis suggests that, in the \ac{SSM}'s regime of operation, increased variability has the effect of reducing the sensitivity of the neuron to other synaptic inputs by flattening the neural transfer curve.
With multiplicative noise, input variability can be high when pre-synaptic neurons with strong synaptic weights are active.
In the \ac{SSM}, such an activity pattern emerges when the probability of a given state under the learned model and the sensory data is high. 
That case suggests that the network has reached a good estimate and should not be easily modified by other evidence, which is the case when the neural transfer curve is flatter. 
Synaptic unreliability can thus play the role of a dynamic normalization mechanism in the neuron with direct implications on probabilistic inference and action selection in the brain. 

Aithchison and Latham suggested the ``synaptic sampling hypothesis'' whereby pre-synaptic spikes would draw samples from a distribution of synaptic weights \cite{Aitchison_Latham13_syna-samp}. 
This process could be a mechanism used in the brain to represent uncertainty over the parameters of a learned model \cite{Aitchison_Latham14_baye-syna}. 
Stochasticity can be carried to the learning dynamics as well. 
Recent studies point to that fact, when learning the blank-out probability at the synapses is also learned \cite{Al-Shedivat_etal15_neur-gene,Al-Shedivat_etal15_inhe-stoc}, the learned neural generative models capture richer representations, especially when labeled data is sparse.
Kappel and colleagues also showed that stochasticity in the learning dynamics improves generalization capability of neural network.

The blank-out noise model used in this work is a particular case of the studies above, whereby the weights of the synapses are either zero or the value of the stored weight with a fixed probability.
In contrast to previous work, we studied the learning dynamics under this probabilistic synapse model within an otherwise deterministic recurrent neural network. 
Besides the remarkable fact that synaptic stochasticity alone is sufficient for sampling, it enables robust learning of sparse representations and an efficient implementation in hardware.

\paragraph{Related Work on Mapping Machine Learned Models onto Neuromorphic Hardware}
Many approaches for configuring spiking neural networks rely on mapping pre-trained artificial neural networks onto spiking neural networks using a firing rate code \cite{OConnor_etal13_real-clas,Diehl_etal15_fast-high,Cao_etal14_spik-deep,Hunsberger_Eliasmith15_spik-deep}.
Many recent work show that the approximations incurred in brain-inspired and neuromorphic hardware platforms \cite{Merolla_etal14_mill-spik,Neftci_etal14_even-cont,Cao_etal14_spik-deep,Diehl_etal15_fast-high,OConnor_etal13_real-clas,Das_etal15_gibb-samp,Marti_etal15_ener-neur} including reduced bit precision \cite{Stromatias_etal15_robu-spik,Muller_Indiveri15_roun-meth,Marti_etal15_ener-neur} have a minor impact on performance.
Standard artificial neural networks such as deep convolutional neural networks and recurrent neural networks trained with Rectified linear units (ReLU) can be mapped on spiking neurons by exploiting the threshold-linear of integrate \& fire neurons \cite{Diehl_etal15_fast-high,Cao_etal14_spik-deep}.
Such mapping techniques have the advantage that they can leverage the capabilities of existing machine learning frameworks such as Caffe \cite{Jia_etal14_caff-conv} or pylearn2 \cite{Goodfellow_etal13_pyle-mach} for brain-inspired computers.
Although mapping techniques do not offer a solution for on-line, real-time learning, they resulted in the best performing spike-based implementations on standard machine learning benchmarks such as MNIST and CIFAR.

\paragraph{Related work on Online Learning with Spiking Neurons}
Training neural networks is a very time- and energy-consuming operation, often requiring multiple days on a computing cluster to produce state-of-the-art results.
Using neuromorphic architectures for learning neural networks can have significant advantages from the perspectives of scalability, power dissipation and real-time interfacing with the environment.
While embedded synapses require additional chip resources and usually prohibits the implementation of more complex rules, a recent survey of software simulators argues that dedicated learning hardware is a prerequisite for real-time learning (or faster) in spiking networks \cite{Zenke_Gerstner14_limi-to}.
This is because the speed-up margin of parallelism encounters a hard boundary due to latencies in the inter-process communications.

The \ac{SSM} is an ideal candidate for \emph{embedded}, online learning, where plasticity is implemented locally using a dedicated on-chip device \cite{Azghadi_etal14_spik-syna}.
These operate on local memory (\emph{e.g.} synaptic weights) using local information (\emph{e.g.} neural events) which will lead to more scalable, faster and more power efficient learning compared to computer clusters, and the ability to adapt in real-time to changing environments.

Previous work reported on on-line learning of MNIST classification with spiking neurons.
A hierarchical spiking neural network with a \ac{STDP}-like learning rule achieved 8\% error rate on the MNIST task \cite{Beyeler_etal13_cate-deci}.
Using models of non-volatile memory devices as synapses and STDP learning, error rates  of 6.3\% were reported \cite{Querlioz_etal15_bioi-prog}. 
Diehl and Cook demonstrated the best results on unsupervised learning with spiking neural networks so far \cite{Diehl_Cook15_unsu-lear}.
Their network model is comparable to competitive learning algorithms where each neuron learns a representation of a portion of the input space.
Their architecture could achieve up to 5\% error rate.
The \ac{SSM} outperformed this best result using a much smaller number of neurons and synapses and relying on dynamics that are more amenable to hardware implementations.
However, the number of repetitions to reach this performance using the \ac{S3M} was larger than the above studies (512,000 presentations vs 40,000 in \cite{Diehl_Cook15_unsu-lear}).
Our neural sampling based architecture with stochastic neurons and deterministic synapses achieved peak performance after 15,000 samples \cite{Neftci_etal14_even-cont}, suggesting that the slowness is partly caused by the stochastic connections. 
Similar results have been observed using the DropConnect algorithm \cite{Wan_etal13_regu-neur}.

\subsection{Implementations of Synaptic Unreliability in Neuromorphic Hardware}
At least four studies reported the implementation of blank-out synapses for neuromorphic systems using the \ac{AER} \cite{Goldberg_etal01_prob-syna,Corradi_etal14_mapp-arbi,Choudhary_etal12_sili-neur,Merolla_etal14_mill-spik}.
In these studies, synaptic unreliability was mainly used as a mechanism for increasing the resolution of the synaptic weights in hardware (which is often binary).
In fact, the mean of a synaptic current produced by a stochastic synapse is the probability times the weight of the synapse.
By allowing the probability to take values with high precision, the effective resolution of the synapse weight can be increased.
The downside is that this approach is valid only when the neural computations are rate-based, such as in the neural engineering framework \cite{Eliasmith_Anderson04_neur-engi} where synaptic unreliability in neuromorphic systems was primarily applied \cite{Corradi_etal14_mapp-arbi,Choudhary_etal12_sili-neur}.

In rate-based models, the variability introduced by stochastic synapses is dealt with by averaging over large populations of neurons or by taking temporal averages.
Implementations based on firing rate codes thus disregard spike times. 
From a hardware perspective, firing rate codes often raise the question whether a spike-based hardware platform is justifiable over a direct, dedicated implementation of the machine learning operations, or even a dedicated implementation of the rate dynamics \cite{Wang_etal15_neur-hard}.
In contrast, codes based on neural sampling, synaptic sampling or phase critically depend on spike statistics or the precise timing of the spikes. 
For example, in the \ac{SSM}, synaptic unreliability and the variability that it causes are an integral part of the sampling process.
The variability introduced by the multiplicative property of synaptic noise is exploited both as a mechanism that generates sigmoidal activation and that improves learning.
Results from the network dynamics suggest that the introduced variability generates sparser representations, and in some cases are insensitive to parameter rescaling.
Thus, our work suggests that synaptic unreliability can play much more active roles in information processing.

The robustness and to synaptic pruning and weight down-sampling is a promising feature to further decrease the hardware footprint of \ac{dSSM}. 
However, these two features are currently introduced post-learning. 
Learning procedures that could introduce synaptic pruning and synaptic generation during (online) learning is the subject of future research.

\pagebreak
\FloatBarrier
\begin{table*}
\begin{center}
    \begin{tabular}{|l | l | c | c |}
    \hline
    $\sigma$        & Noise amplitude                                               & \ac{S3M}, visible neurons               & $\unit[4.47]{nA}$ \\
    $p$             & Blank-out probability at synapse                              & all models                                 & $\unit[.5]{}$ \\
    $\tau_{r}$      & Refractory period                                             & all \ac{S3M}                                & $\unit[4]{ms}$\\
    $\tau_{syn}$    & Time constant of recurrent, and bias synapses.                & all \ac{S3M}                                & $\unit[4]{ms}$\\
    $\tau_{br}$     & ``Burn-in'' time of the neural sampling                       & all \ac{S3M}                                & $\unit[10]{ms}$\\
    $g_{L}$         & Leak conductance                                              & all \ac{S3M}                                & $\unit[1]{nS}$ \\
    $u_{rst}$       & Reset Potential                                               & all \ac{S3M}                                & $\unit[0]{V}$ \\
    $C$             & Membrane capacitance                                          & all \ac{S3M}                                & $\unit[1]{pF}$ \\
    $\theta$        & Firing threshold                                              & all \ac{S3M}                                & $\unit[100]{mV}$ \\
    $2T$            & Epoch duration                                                & all \ac{S3M}                                & $\unit[100]{ms}$\\
    $W$             & Initial weight matrix                                         & all \ac{S3M}                                & $N(0, .3)$ \\
    &                                                                               & all \ac{dSSM}, RBM                          & $N(0, .1)$ \\
    $b_v, b_h$      & Initial bias for layer $v$ and $h$                            & all \ac{S3M}                                & $-.15$ \\
                    &                                                               & all \ac{dSSM}, RBM                          & $0$ \\
    $T_{sim}$       & Simulation time per epoch                                     & all \ac{S3M}                                & $\unit[100]{s}$ \\
    $N_v, N_h$      & Number of visible and hidden units                            & all models,                                & $794, 500$ \\
                    &                                                               & except in \reffig{fig:kl_divergence_bssm}  & $5, 5$ \\
    $N_c$           & Number of class label units                                   & all models                                 & $10$ \\
    $N_{samples}$   & Total number of MNIST sample presentations                    & \reffig{fig:fig_convergence}, \ac{S3M}  & $\unit[512000]{}$ \\
                    &                                                               & \reffig{fig:fig_convergence}, \ac{dSSM}, RBM    & $\unit[75\cdot10^{6}]{}$ \\
    $\tau_{STDP}$   & Learning time window                                          & all models                                 & $\unit[10]{ms}$ \\
    $\epsilon_{q}$  & Learning rate for $W$                                         & all \ac{S3M}                                & $\unit[3.85\cdot 10^{-6}]{}$ \\
    $\epsilon    $  &                                                               & all \ac{dSSM}, RBM                                  & $.025$ \\
    $\epsilon_{b}$  & Learning rate for $b_v$, $b_h$                                & all \ac{S3M}                                & $\unit[1.43\cdot 10^{-5}]{}$ \\
                    &                                                               & all \ac{dSSM}, RBM                                  & $.025$ \\
    $n_{batch}$     & Batch size                                                    & all \ac{dSSM}, RBM                                  & $50$\\
    $W$             & Distribution of weight parameters                             & \reffig{fig:kl_divergence_bssm}      & $N(-.3,1.5)$\\
    $b_{h},b_{v}$   & Distribution of bias   parameters                             & \reffig{fig:kl_divergence_bssm}      & $N(0,1.5)$\\
    \hline
  \end{tabular}
\end{center}
\caption{\label{tab:parameters} Parameters used for the neural networks.}
\end{table*}

\FloatBarrier
\section{Acknowledgments}
We thank Friedemenn Zenke for support on the Auryn simulator and discussion.\\
This work funded by the National Science Foundation (NSF CCF-1317373, EN, BP, SJ, GC), the Office of Naval Research (ONR MURI 14-13-1-0205, EN, BP), and Intel Corporation (EN, GC). 


\end{document}